%% file: main.tex
\documentclass[10pt,twocolumn,letterpaper]{article}

\usepackage[pagenumbers]{cvpr} %

\definecolor{cvprblue}{rgb}{0.21,0.49,0.74}
\usepackage[breaklinks,colorlinks,allcolors=cvprblue]{hyperref}

\input{preamble}

\def\paperID{44851} %
\def\confName{CVPR}
\def\confYear{2026}

\title{\modelname: Language-driven Dexterous Grasp Generation with Embodied Reasoning}

\author{
Junha Lee$^{1}$\qquad\qquad
Eunha Park$^{1}$\qquad\qquad
Minsu Cho$^{1,2}$\vspace{1mm}
\\ 
{$^{1}$Pohang University of Science and Technology (POSTECH)\qquad$^{2}$RLWRLD} \vspace{1mm}
\\ 
\small\url{https://junha-l.github.io/dexter}
}

\begin{document}
\twocolumn[{%
    \maketitle
    \renewcommand\twocolumn[1][]{#1}%
    \vspace{-5mm} 
    \centering
    \includegraphics[width=.999\linewidth]{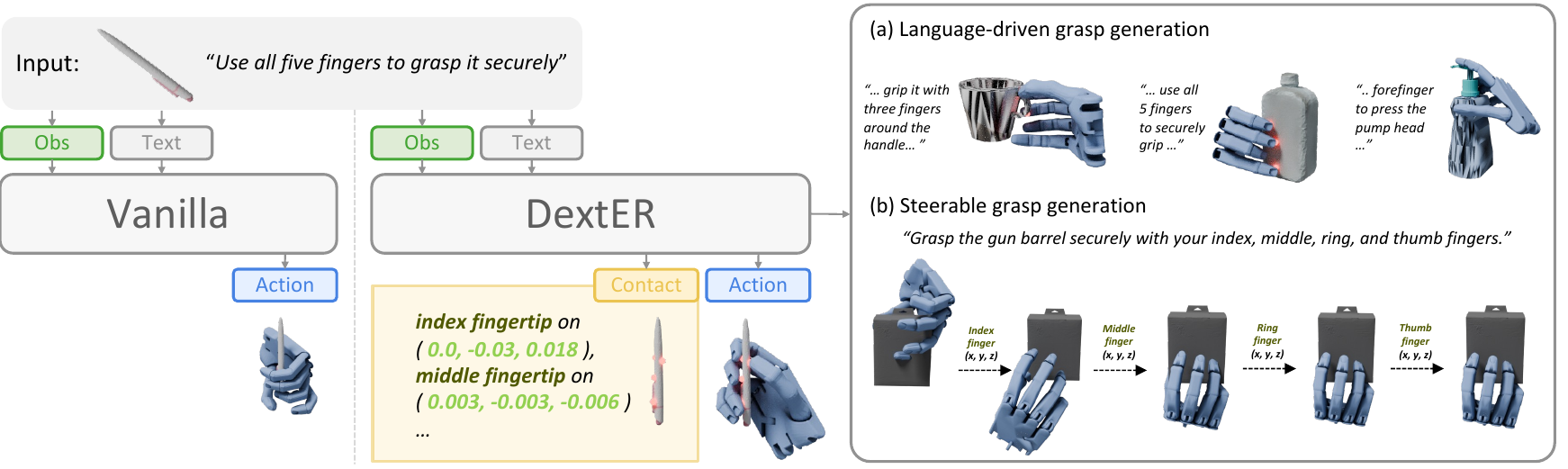}
    \captionof{figure}{
    \textbf{\modelname}
        introduces contact-based embodied reasoning for language-driven dexterous grasp generation.
        Given a 3D object and instruction, \modelname autoregressively predicts which finger links contact where on the object surface before generating the final grasp.
        Our method achieves state-of-the-art performance with significant improvement in intention alignment and enables steerable generation where users can guide grasp synthesis by specifying partial contact constraints.
    }
    \vspace{4mm}
}]

\input{sec/0_abstract}    
\input{sec/1_intro}
\input{sec/2_related}
\input{sec/3_method}
\input{sec/4_experiment}

\input{sec/5_conclusion}

\clearpage
\noindent \textbf{Acknowledgements.}
This work was supported by the IITP grants (RS-2022-II220290: Visual Intelligence for Space-Time Understanding and Generation based on Multi-layered Visual Common Sense (50\%), RS-2024-00457882: National AI Research Lab Project (45\%), RS-2019-II191906: AI Graduate School at POSTECH (5\%))
funded by the Korea government (MSIT).

{
    \small
    \bibliographystyle{ieeenat_fullname}
    \bibliography{main}
}

\appendix
\input{sec/X_suppl}

\end{document}

%% file: preamble.tex
\usepackage[utf8]{inputenc} %
\usepackage[T1]{fontenc}    %
\usepackage{hyperref}       %
\usepackage{url}            %
\usepackage{booktabs}       %
\usepackage{amsfonts}       %
\usepackage{nicefrac}       %
\usepackage{microtype}      %
\usepackage{xcolor}         %
\usepackage{colortbl}
\usepackage{graphicx}
\usepackage{multirow}
\usepackage{makecell}
\usepackage{tablefootnote}
\usepackage{amsmath}
\usepackage{amssymb}
\usepackage{pifont}         %
\usepackage{xspace}
\usepackage{wrapfig}
\usepackage{caption}
\usepackage{algorithm}
\usepackage{dsfont}
\usepackage{algpseudocode}
\usepackage{algorithmicx}
\usepackage{pgfplots}
\usepackage{fancyvrb}
\usepackage{tikz}
\usetikzlibrary{patterns,decorations.pathreplacing}
\pgfplotsset{compat=newest}

\newcommand{\TODO}[1]{\textbf{\color{red}[TODO: #1]}}
\renewcommand{\TODO}[1]{}

\newcommand{\modelname}{DextER\xspace}

\newcommand{\Sref}[1]{Sec.~\ref{#1}}

\newcommand{\noindentbold}[1]{\noindent \textbf{#1.}\xspace}

\newcommand{\prompttoken}[2]{\colorbox{#1}{\texttt{#2}}}
\newcommand{\ntoken}{\colorbox{gray!10}{\texttt{\char`\\n}}}
\newcommand{\jointtoken}[1]{\prompttoken{orange!15}{#1}}
\newcommand{\actiontoken}[1]{\prompttoken{purple!15}{#1}}
\newcommand{\postoken}[1]{\prompttoken{cyan!15}{#1}}

%% file: sec/0_abstract.tex
\begin{abstract}
Language-driven dexterous grasp generation requires the models to understand task semantics, 3D geometry, and complex hand-object interactions.
While vision-language models have been applied to this problem, existing approaches directly map observations to grasp parameters without intermediate reasoning about physical interactions.
We present \modelname, \textbf{Dex}terous Grasp Generation with \textbf{E}mbodied \textbf{R}easoning, which introduces contact-based embodied reasoning for multi-finger manipulation.
Our key insight is that predicting which hand links contact where on the object surface provides an embodiment-aware intermediate representation, bridging task semantics with physical constraints.
\modelname autoregressively generates embodied contact tokens specifying which finger links contact where on the object surface, followed by grasp tokens encoding the hand configuration.
On DexGYS, \modelname achieves 67.14\% success rate, outperforming state-of-the-art by 3.83 p.p. with 96.4\% improvement in intention alignment.
We also demonstrate steerable generation through partial contact specification, providing fine-grained control over grasp synthesis.
\end{abstract}

%% file: sec/1_intro.tex
\section{Introduction}
\label{sec:intro}

Dexterous manipulation with multi-fingered robotic hands remains one of the most challenging problems in robotics, requiring precise coordination to achieve stable, task-appropriate grasps.
With anthropomorphic hands possessing 20+ degrees of freedom, generating such grasps extends far beyond the capabilities of parallel-jaw grippers, demanding reasoning about intricate contact patterns, finger coordination, and task-specific constraints.
This challenge is further amplified as robots increasingly operate in unstructured environments alongside humans, where language-driven dexterous control becomes crucial for flexible, general-purpose robotic manipulation.

To address these challenges, recent approaches~\cite{he2025dexvlg,li2024semgrasp,wei2024grasp} employ vision-language models (VLMs) that fuse 3D visual representations with linguistic understanding, enabling dexterous grasp generation through multimodal understanding of object geometry and language instructions.
While effective, these models typically map observations directly to grasp configurations without explicitly modeling intermediate physical interactions. This approach can miss important structural priors about how multi-fingered hands interact with objects, limiting grasp quality and intention alignment.
To incorporate such structural reasoning, chain-of-thought (CoT) approaches~\cite{wei2022chain} have been extended to robotics through embodied reasoning~\cite{clark2025action,huang2025thinkact,sun2025emma,zawalski2025robotic,chen2025training,zhao2025cot}, where models generate intermediate steps before predicting actions, such as textual plans describing task strategies, object descriptions capturing visual semantics, or spatial references like bounding boxes localizing relevant regions.
However, applying this reasoning approach to dexterous grasping faces a unique challenge: what intermediate representation captures the physical interaction structure critical for multi-finger manipulation?

We propose \textbf{\modelname}, a vision-language framework that addresses this challenge through \textit{contact-based embodied reasoning} for dexterous grasp generation.
Our key insight is that predicting which hand links make contact and where on the object surface provides an embodiment-aware intermediate representation that bridges high-level task semantics (e.g., ``grasp the mug by its handle'') with physical constraints of both the robot embodiment and object geometry.
Unlike existing approaches that directly predict grasp parameters, \modelname autoregressively generates grasps through structured contact reasoning:
The model first produces \textit{embodied contact tokens} that specify which finger links contact the object and their 3D positions on the object surface, followed by \textit{grasp tokens} that encode the complete hand configuration.
All tokens are generated autoregressively within a unified next-token prediction framework, enabling interpretable contact reasoning as an intermediate step.

Our key contributions are threefold.
\textbf{First}, we propose contact-centric reasoning as an effective embodied thinking process for language-driven dexterous grasp generation, achieving state-of-the-art performance by explicitly modeling the physical interaction structure between multi-finge
red hands and objects.
\textbf{Second}, we annotate DexGYS and Dexonomy datasets with physics-based contact annotations using MuJoCo simulator and natural language grasp descriptions using VLM, enabling large-scale training of contact-aware models.
\textbf{Third}, our autoregressive generation framework enables \textit{steerable grasp generation}, where users can guide the model by specifying partial contact constraints (e.g., a few finger links in contact with the object and their 3D coordinates), and the model completes the remaining sequence while respecting these constraints, providing an effective interface for fine-grained control over grasp synthesis.

Through extensive experiments, we show that \modelname achieves a 67.14\% grasp success rate on the DexGYS benchmark, outperforming prior state-of-the-art methods by 3.83 percentage points.
Our model demonstrates 96.4\% improvement in intention alignment, with higher diversity in generated grasp configurations.
The explicit contact predictions provide insights into the model's reasoning process, with qualitative analysis revealing physically meaningful contact patterns that align with human grasping strategies.
Additionally, we demonstrate that partial contact specifications enable steerable grasp generation, providing fine-grained user control while maintaining grasp quality.

%% file: sec/2_related.tex
\input{figures/02_architecture/figure}

\section{Related Work}
\label{sec:related_work}

\noindentbold{Dexterous grasp generation}
Early approaches to dexterous grasp generation relied on analytic optimization of grasp quality metrics such as force closure and wrench resistance~\cite{liu2021synthesizing, li2022gendexgrasp,wang2023dexgraspnet,zhang2024dexgraspnet,chen2025bodex,chen2025dexonomy} or differentiable simulation~\cite{turpin2022grasp, turpin2023fast}.
However, these methods require expensive physics-based optimization at test time and struggle to generalize beyond known object geometries.
The emergence of large-scale grasp datasets~\cite{xu2023unidexgrasp,wang2023dexgraspnet,zhang2024dexgraspnet,chen2025bodex} enabled a paradigm shift toward data-driven learning, where models predict stable hand configurations directly from visual observations.
Building on this foundation, recent works employ generative modeling frameworks, including diffusion models, transformers, and reinforcement learning policies, to capture the multimodal distribution of stable grasps~\cite{huang2023diffusion, xu2023unidexgrasp, wan2023unidexgrasp++, zhang2024graspxl, lu2024ugg, weng2024dexdiffuser,xu2024dexterous,zhong2025dexgrasp}.
Despite achieving impressive success rates on physically stable grasp generation, these methods operate purely on visual input and lack the ability to incorporate task semantics or functional requirements.
This is a critical limitation for real-world manipulation where grasp intent determines success.

\noindentbold{Language-conditioned dexterous grasp generation}
To enable task-specific functional grasping~\cite{agarwal2023dexterous,wu2024cross,huang2025fungrasp}, recent work has integrated language conditioning into dexterous grasp generation.
These methods follow two distinct paradigms, each with complementary limitations.
\textit{Two-stage pipelines} decompose the problem into separate intention understanding and grasp synthesis modules~\cite{wei2025afforddexgrasp,zhong2025dexgraspvla,wei2024grasp}.
These approaches first use vision-language models to identify task-relevant object regions or affordances from language instructions, then feed these intermediate representations to separate grasp generators.
While this modular design provides interpretable intermediate outputs, the disjoint training prevents mutual learning between semantic understanding and physical grasp synthesis.
In contrast, \textit{end-to-end approaches} learn direct mappings from multimodal inputs to grasp parameters through joint training~\cite{li2024multi, li2024semgrasp, he2025dexvlg}.
This enables faster inference and implicit alignment between vision, language, and actions.
However, they lack explicit intermediate reasoning about physical interactions, making it difficult to interpret failures or adapt to novel task specifications.
Critically, both paradigms fail to explicitly model the fundamental physical principle that successful grasps depend on where and how the hand makes contact with objects.

\noindentbold{Vision-Language-Action models}
General VLA frameworks have demonstrated impressive capabilities by transferring the rich common sense and world knowledge learned by vision-language models to robotic action tasks~\cite{brohan2022rt,zitkovich2023rt,driess2023palm,black2024pi,bjorck2025gr00t,kim2025openvla,deng2025graspvla}.
By unifying perception, instruction understanding, and control within transformer-based policies, these models can leverage large-scale pretraining on diverse robot demonstrations and web data.
While achieving strong generalization on manipulation tasks with parallel-jaw grippers, they lack specialized designs for high-dimensional dexterous control and operate primarily as end-to-end perception-to-action mappings without explicit physical reasoning.

\noindentbold{Embodied Chain-of-Thought reasoning}
Recent work has begun incorporating explicit reasoning steps between perception and action to improve interpretability and generalization.
These approaches train policies to generate intermediate representations before executing final actions~\cite{clark2025action, huang2025thinkact, sun2025emma, deng2025graspvla,lee2025molmoact,zhao2025cot,chen2025training,zawalski2025robotic}.
Such representations include textual plans, bounding boxes or points, trajectory waypoints, or latent reasoning traces.
By decomposing complex tasks into interpretable reasoning steps, these models demonstrate improved sample efficiency, generalization, and explainability on mobile manipulation and parallel-jaw grasping tasks.
However, their application to dexterous multi-finger manipulation remains unexplored.
The key challenge lies in designing intermediate representations that capture the unique constraints of multi-finger grasping, including complex contact geometry, joint coordination, and the mapping between task semantics and physical interactions.

We bridge these gaps by introducing contact-centric reasoning as an embodied chain-of-thought process specifically designed for dexterous manipulation.
Our model explicitly predicts where the hand should make contact with objects before generating grasp actions.
This design unifies semantic understanding and physical grasp synthesis while maintaining interpretability through explicit contact predictions, addressing the limitations of previous approaches.

%% file: figures/02_architecture/figure.tex
\begin{figure*}[!t]
    \centering
    \includegraphics[width=\linewidth]{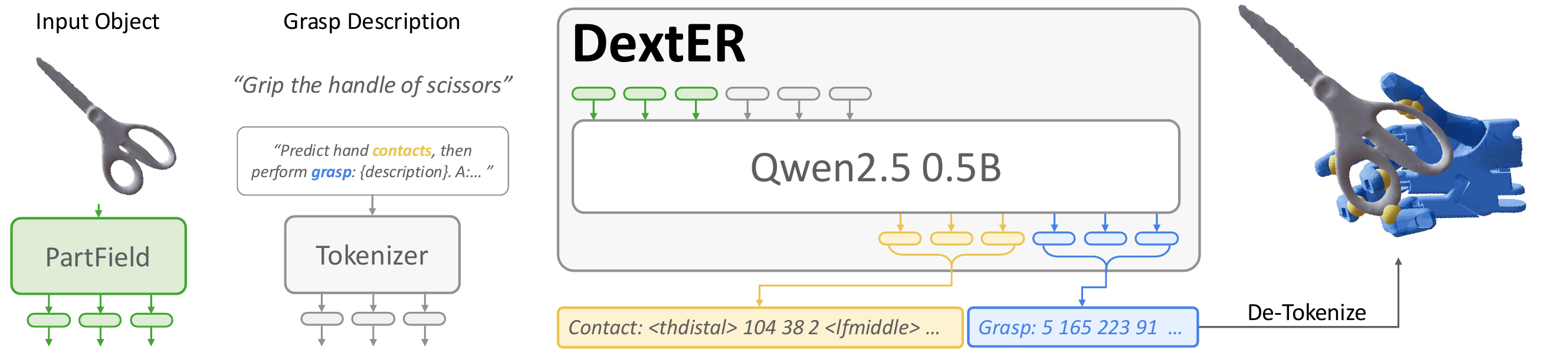}
    \caption{\textbf{\modelname model architecture.}
    Our model processes 3D point clouds and language instructions to predict dexterous grasping actions for the multi-fingered robotic hand.
    (Left) The input point clouds and textual grasp descriptions are encoded into tokens using a pretrained point cloud encoder~\cite{liu2025partfield} and a text tokenizer~\cite{qwen2,qwen2.5}.
    (Middle) The LLM backbone~\cite{qwen2,qwen2.5} fuses point cloud embeddings with text prompts and autoregressively generates discretized contact and action tokens.
    (Right) The generated contact and action tokens are de-tokenized into contact positions, hand joint configurations, and grasp poses.
    }
    \label{fig:architecture}
\end{figure*}

%% file: sec/3_method.tex
\section{Method}
\label{sec:method}
Given a point cloud $\mathbf{P} \in \mathbb{R}^{N \times 3}$ of the target object and a language instruction $\mathbf{T}$ describing the desired grasp, our goal is to predict a grasp pose $\mathbf{a} \in \mathbb{R}^{D}$ for a multi-fingered hand, where $D$ denotes the hand's degrees of freedom.
Rather than directly mapping inputs to grasp parameters, \modelname factorizes this prediction through contact patterns $\mathcal{C}$ as an intermediate reasoning step:
\begin{equation}
p(\mathbf{a}, \mathcal{C} | \mathbf{P}, \mathbf{T}) = p(\mathcal{C} | \mathbf{P}, \mathbf{T}) \cdot p(\mathbf{a} | \mathcal{C}, \mathbf{P}, \mathbf{T}),
\end{equation}
where the model first predicts which finger links contact the object and where, then generates the full grasp configuration conditioned on these contacts.
This section describes the model architecture (Sec.~\ref{subsec:model}), embodied reasoning formulation (Sec.~\ref{subsec:embodied_cot}), training strategy (Sec.~\ref{subsec:training}), and dataset curation (Sec.~\ref{subsec:dataset}).

\subsection{Model Architecture}
\label{subsec:model}

Following recent vision-language-action (VLA) models, we extend a vision-language model to predict robot actions.
Among various design choices for action representation, we discretize continuous grasp parameters into discrete tokens~\cite{zitkovich2023rt,kim2025openvla,pertsch2025fast,lee2025molmoact} and autoregressively generate them using an LLM backbone.
The overview of the model can be found in Fig.~\ref{fig:architecture}.
The model comprises three components: a 3D vision encoder, a multimodal projector, and an LLM backbone.
The vision encoder extracts geometric representations from point clouds, which are aligned to the language model's embedding space via a lightweight projector.
The projected visual tokens are concatenated with language tokens and fed into the LLM backbone, enabling multimodal understanding and autoregressive generation of discrete grasp tokens.

\noindentbold{Point Cloud Encoding}
Given an input point cloud $\mathbf{P} \in \mathbb{R}^{N \times 3}$ of the target object, we extract geometric features $\mathbf{F} \in \mathbb{R}^{M \times d}$ using a pretrained 3D encoder, where $M$ and $d$ denote the number of tokens and feature dimension.
We adopt PartField~\cite{liu2025partfield} as our 3D encoder, which is pretrained for part-segmentation via contrastive learning with 2D SAM masks, yielding part geometry-aware features.
Its part-level spatial understanding is particularly suited for contact reasoning, as predicting finger-object contacts requires precise localization on the object surface.
We project the features to the LLM's embedding space via a lightweight MLP.

\noindentbold{Action Tokenization}
To enable autoregressive generation in the discrete token space, we tokenize continuous robot actions into discrete bins. 
For a grasp action $\mathbf{a} \in \mathbb{R}^{D}$, where it includes 6D pose of the wrist and joint angles of the fingers, we uniformly bin each dimension into $N_{\mathbf{a}}$ bins and represent each binned value as a unique token $\langle\text{action\_bin}\_i\rangle$. 
The complete action sequence is wrapped with special tokens: $\langle|{\text{action\_start}}|\rangle \, \{\text{action tokens}\} \, \langle|{\text{action\_end}}|\rangle$. 
This design allows us to leverage the VLM's next-token prediction objective for action generation while maintaining compatibility with the pretrained tokenizer vocabulary.

\subsection{Embodied Reasoning via Contact Prediction}
\label{subsec:embodied_cot}

Pure end-to-end learning from vision to actions often lacks interpretability and fails to capture explicit physical reasoning about manipulation tasks.
We hypothesize that predicting contact points where the hand will touch the object serves as an effective form of embodied reasoning, as successful grasps fundamentally depend on appropriate contact configurations.
This intermediate prediction provides an interpretable representation of the model's understanding of object affordances and grasp requirements.

\noindentbold{Meta-prompts for eliciting contact reasoning}
To guide the model to engage in contact reasoning before action generation, we employ meta-prompts that explicitly instruct the model to predict contacts.
These meta-prompts are prepended to the task instruction and encourage step-by-step reasoning, such as ``Think step by step: first predict which links contact where on the object, then predict the grasp pose'' or ``Predict the contact points between the hand and object, then generate the grasp configuration''.
During training, we use a diverse set of meta-prompt variations to prevent overfitting to specific phrasings and encourage robust reasoning capabilities.

\noindentbold{Contact representation}
We represent contacts as a set of link-position pairs $\mathcal{C} = \{(l_i, \mathbf{p}_i)\}$, where $l_i$ denotes a robotic hand link (e.g., index finger middle link) and $\mathbf{p}_i \in \mathbb{R}^3$ is the 3D contact position on the object surface.
Following the action tokenization approach, we first normalize coordinates of contact positions into a fixed 3D bounding box computed from the dataset, then uniformly discretize each spatial dimension into $N_{\text{pos}}$ bins and map them to position tokens.
Link names are encoded as link tokens $\langle l_i \rangle$ (e.g., $\langle$\texttt{thbase}$\rangle$, $\langle$\texttt{ffdistal}$\rangle$, $\langle$\texttt{mfmiddle}$\rangle$ for thumb base, first finger distal, and middle finger middle links).
Each contact is represented as a sequence $\langle l_i \rangle \langle p_{ix} \rangle \langle p_{iy} \rangle \langle p_{iz} \rangle$, where $\langle p_{ix} \rangle$, $\langle p_{iy} \rangle$, $\langle p_{iz} \rangle$ are the discretized position bin tokens for the x, y, z coordinates of $\mathbf{p}_i$.
For example, a thumb contact might be encoded as $\langle$\texttt{thbase}$\rangle \langle p_{x} \rangle \langle p_{y} \rangle \langle p_{z} \rangle$.
The full contact prediction is wrapped with special delimiters: $\langle|{\text{contact\_start}}|\rangle \, \{\text{contact tokens}\} \, \langle|{\text{contact\_end}}|\rangle$.

To support these representations, we register all required tokens as special tokens in the pretrained tokenizer: action bin tokens ($N_{\mathbf{a}}$ tokens), position bin tokens ($N_{\text{pos}}$ tokens), link tokens for all ShadowHand links (e.g., $\langle$\texttt{thbase}$\rangle$, $\langle$\texttt{ffdistal}$\rangle$), and delimiter tokens.
This expands the tokenizer vocabulary while preserving the pretrained model's language understanding capabilities.

\subsection{Training Strategy}
\label{subsec:training}

We train the model end-to-end using standard next-token prediction over the full sequence: point cloud tokens, task description, contact tokens, and action tokens.
The model learns to first predict contact patterns, then generate corresponding grasp poses autoregressively.

\noindentbold{Hybrid attention mechanism}
We employ a hybrid attention pattern where point cloud tokens use bidirectional attention to capture global geometric context, while language and action tokens use causal attention.
This allows point cloud features to interact fully for comprehensive 3D understanding, while maintaining standard autoregressive generation for text and action sequences.

\noindentbold{Contact position dropout}
For regularization, we introduce contact position dropout during training.
With probability $p_{\text{drop}}$, we remove the position tokens $\langle p_{ix} \rangle \langle p_{iy} \rangle \langle p_{iz} \rangle$ from contact sequences while retaining link tokens $\langle l_i \rangle$.
This creates training samples where the model sees only which links will contact (e.g., $\langle$\texttt{thbase}$\rangle \langle$\texttt{ffdistal}$\rangle$) without specific positions.
This dropout prevents overfitting to specific token patterns and enables the model to reason from varying levels of contact detail during training and inference.

\subsection{Dataset Curation}
\label{subsec:dataset}

\noindentbold{Source datasets}
We curate training data from two large-scale dexterous grasp datasets: DexGYS~\cite{wei2024grasp} and Dexonomy~\cite{chen2025dexonomy}.
DexGYS provides diverse grasp poses across thousands of objects with language descriptions, offering broad coverage of geometries and functional scenarios.
Dexonomy organizes grasps into a taxonomy of 31 types (e.g., power grasp, precision pinch), providing systematic coverage of hand configurations and contact patterns.
We leverage their complementary strengths: DexGYS for scale and language, Dexonomy for structured grasp variations.

\noindentbold{Physics-based contact annotation}
To enable the proposed embodied reasoning, we augment both datasets with structured contact annotations.
Since manual contact annotation is prohibitively expensive, we automatically generate structured contact data using MuJoCo's physics engine.
For each grasp, we load hand and object models into MuJoCo, execute forward kinematics, and extract contacts from the physics buffer.
This produces paired data: (1) \textit{contact anatomy}, identifying which finger links make contact, and (2) \textit{contact positions}, specifying 3D locations on the object surface.
We apply this pipeline to both datasets.

\noindentbold{Grasp instruction annotation}
While Dexonomy's structured taxonomy enables systematic grasp reasoning, it lacks language descriptions unlike DexGYS.
We develop an automatic pipeline to generate grasp descriptions for Dexonomy using a vision-language model~\cite{team2025gemma}.
For each grasp, we render five multi-view images and prompt the VLM conditioned on both the renderings and contact anatomy.
The VLM identifies the object category, infers contacted functional parts (e.g., handle, rim), and generates textual grasp descriptions.

%% file: sec/4_experiment.tex
\section{Experiments}
\label{sec:experiment}

In this section, we evaluate \modelname on language-driven dexterous grasp generation tasks.
We first outline the implementation details (Sec.~\ref{subsec:impl_details}), followed by evaluation on language-conditioned grasp generation (Sec.~\ref{subsec:language_grasp}), ablation studies (Sec.~\ref{subsec:ablation}), cross-dataset generalization (Sec.~\ref{subsec:zeroshot}), and steerable generation (Sec.~\ref{subsec:steerable}). 

\subsection{Implementation Details}
\label{subsec:impl_details}

\noindentbold{Model architecture}
We use the pretrained PartField~\cite{liu2025partfield} as our visual encoder and initialize the language backbone from Qwen2.5-0.5B~\cite{qwen2.5,qwen2}, the smallest model in the Qwen2.5 family.
For efficiency, we downsample the triplane feature maps at the PartField encoder bottleneck, resulting in 768 visual tokens (\eg 3 $\times$ 16 $\times$ 16).
The multimodal projector is implemented as a 2-layer MLP.
We tokenize continuous grasp parameters into discrete tokens by first applying quantile normalization to map values from their 1st and 99th percentiles to $[-1, 1]$, then uniformly binning each dimension into $N_{\mathbf{a}}=256$ bins.
For contact positions, we compute the min \& max bounds on the training set and set slightly larger bounds for each dataset, then uniformly discretize into $N_{\text{pos}}=256$ bins per dimension. 

\noindentbold{Training details}
We train \modelname with a batch size of 64 for 100K iterations using the AdamW optimizer with a learning rate of 1e-4 and cosine decay schedule.
We employ mixed-precision (bfloat16) training and gradient clipping at 1.0 to stabilize optimization.
All experiments are implemented in PyTorch and trained on 8 NVIDIA A6000 GPUs.
Training takes approximately 48 hours for 100K iterations.

\noindentbold{Simulation environment}
For the DexGYS dataset, we use Isaac Gym-based simulation following Grasp-as-You-Say~\cite{wei2024grasp} and DexGraspNet~\cite{wang2023dexgraspnet} for grasp execution and stability measurement.
For Dexonomy experiments, we utilize DexGraspBench~\cite{chen2025bodex}, a MuJoCo-based dexterous grasp simulator benchmark. 

\subsection{DexGYS Benchmark}
\label{subsec:language_grasp}

We evaluate \modelname on language-conditioned dexterous grasp generation using the DexGYS~\cite{wei2024grasp} validation split.
Each test sample consists of a 3D object point cloud and a natural language instruction specifying the intended grasp, such as \textit{``grasp the handle to pour''} or \textit{``hold the lid to open''}. The test set comprises entirely unseen objects.

\noindentbold{Evaluation metrics}
Following \cite{wei2024grasp}, we evaluate generated grasps across three aspects: intention consistency (P-FID, Chamfer Distance, Contact Distance), physical stability (Success Rate, $Q_1$, Penetration), and diversity ($\delta_t$, $\delta_r$, $\delta_q$).
Detailed definitions of each metric are provided in Sec.~\ref{subsec:eval_metrics}.

\noindentbold{Results}
\input{tables/01_dexgys_eval}
Table~\ref{tab:dexgys_eval} presents quantitative comparison against baseline methods.
\modelname significantly outperforms all baselines across most metrics.
Most notably, our model achieves the best P-FID score of 0.20, representing a 96.4\% improvement over the previous state-of-the-art DexGYSNet~\cite{wei2024grasp}.
This substantial gain indicates much better alignment between generated grasps and task intentions specified in language.
Note that while CD measures nearest-neighbor proximity to any valid ground-truth grasp, P-FID evaluates distributional similarity and penalizes missing modes.
\modelname achieves ${\sim}2\times$ higher diversity ($\delta_q{=}13.63$ vs.\ 6.12), confirming broader mode coverage rather than collapse onto dense GT modes.
Our approach also achieves the highest success rate at 67.14\%, demonstrating strong performance in generating physically stable grasps.

To understand the role of embodied reasoning, we compare our full model with a variant that directly predicts grasps without intermediate contact prediction (w/o ER).
This comparison reveals that embodied reasoning is critical for both intention alignment and physical quality: without explicit contact prediction, P-FID degrades from 0.20 to 0.30 (50\% increase), and the success rate drops from 67.14\% to 62.37\%.
The variant also exhibits lower force-closure quality ($Q_1$ decreases from 0.89 to 0.66) and increased penetration (0.44 vs. 0.37).
These results suggest that explicit contact reasoning not only helps understand task-specific grasp intentions but also guides the model toward physically stable and high-quality grasps.

Beyond accuracy, \modelname captures the multimodal nature of valid grasps more effectively than prior approaches.
Our autoregressive generation framework produces substantially more diverse solutions, particularly in rotation space where we observe the highest variation.
This diversity is crucial for real-world deployment where multiple valid grasp strategies may exist for a given task instruction.
Fig.~\ref{fig:qual} shows qualitative comparisons on DexGYS validation set.

\subsection{Ablation Study}
\label{subsec:ablation}

To understand the contribution of key design choices, we conduct systematic ablation studies on the DexGYS dataset.
Table~\ref{tab:ablation} presents comprehensive results across six dimensions: embodied chain-of-thought reasoning, action token discretization, position token discretization, contact position dropout, point cloud encoder architecture, and LLM backbone.

\noindentbold{Embodied Chain-of-Thought (ECoT)}
We first evaluate the impact of our embodied chain-of-thought reasoning mechanism, which predicts contact positions before generating grasp actions.
The results demonstrate the critical importance of this explicit reasoning step: removing ECoT causes P-FID to degrade from 0.20 to 0.30 (50\% increase), while the success rate drops from 67.14\% to 62.37\%.
This significant performance gap confirms that decomposing the grasp generation task into contact reasoning followed by action synthesis leads to both better intention alignment and higher physical quality.
The explicit contact prediction provides an interpretable intermediate representation that guides the subsequent action generation.

\noindentbold{Token discretization granularity}
We evaluate the impact of discretization granularity for both action tokens ($N_\mathbf{a}$) and position tokens ($N_{\text{pos}}$), varying the number of bins from 128 to 512.
For both token types, 256 bins consistently achieves optimal performance ($\text{P-FID}=0.20, \text{CD}=1.46, \text{Success}=67.14\%$).
Coarser discretization (128 bins) slightly reduces precision, while finer discretization (512 bins) degrades performance due to increased vocabulary complexity, particularly for actions where P-FID increases to 0.26.
This sweet spot at 256 bins balances expressiveness with learnability, providing sufficient resolution for both precise grasp and contact locations.

\noindentbold{Contact position dropout}
We study the effect of contact position dropout probability $p_{\text{drop}}$ on grasp generation quality by varying it from $p_{\text{drop}}=0.0$ (no dropout) to 1.0 (always drop).
Results show that $p_{\text{drop}}=0.5$ achieves best performance with $\text{P-FID}=0.20$ and $\text{consistency}=0.34$.
Without dropout ($p_{\text{drop}}=0.0$), the model shows slightly worse generalization ($\text{P-FID}=0.22$).
Higher dropout rates progressively degrade performance, with $p_{\text{drop}}=1.0$ essentially removing the benefit of contact position reasoning, approaching the performance of the model without ECoT.
This validates moderate dropout provides effective regularization, improving the model's ability to handle varying levels of contact specification.

\noindentbold{Point cloud encoder}
We evaluate the impact of the point cloud encoder architecture by comparing Uni3D~\cite{zhou2024uni3d} and PartField~\cite{liu2025partfield}.
PartField achieves the best performance ($\text{P-FID}=0.20$, $\text{Success}=67.14\%$) due to its part-aware feature extraction, which naturally aligns with our contact-based reasoning that predicts link-position pairs.
The part-centric representations enable better localization of contact positions on object surfaces, helping the model predict which hand links contact where on functional object regions.

\noindentbold{LLM backbone}
We evaluate the impact of LLM backbone choice by comparing Qwen2.5-0.5B (default) with a 3$\times$ larger Qwen2.5-1.5B~\cite{qwen2.5} and SmolLM2-360M~\cite{allal2025smollm2} (a LLaMA-architecture model).
Scaling to Qwen2.5-1.5B provides only modest improvements (Success: 67.55\% vs.\ 67.14\%), while SmolLM2-360M achieves comparable results (Success: 64.87\%), confirming that performance gains stem primarily from our contact-centric reasoning rather than LLM scale or architecture choice.

\input{figures/03_qual/figure}
\input{tables/03_ablation}

\subsection{Dexonomy Benchmark}
\label{subsec:zeroshot}

To further measure the robustness and generalization ability of the proposed model, we conduct zero-shot generalization experiments. 
In this experiment, we use Dexonomy~\cite{chen2025dexonomy} as training and evaluation dataset. 
Specifically, we carefully designed the data split to test the model's ability to generalize to novel objects and grasp taxonomies.

\noindentbold{Data split}
To evaluate generalization capabilities, we construct one training split and four validation splits based on object and grasp taxonomy overlap with the training set.
\textit{Seen Objects \& Taxonomy} contains objects and grasp taxonomies present in training.
\textit{Unseen Objects} introduces novel objects with familiar grasp taxonomies, while \textit{Unseen Grasp Taxonomy} uses familiar objects with novel grasp taxonomies.
\textit{Unseen Both} tests on both novel objects and grasp taxonomies.
The detailed object categories and grasp taxonomy splits can be found in Sec.~\ref{subsec:eval_metrics}.

\noindentbold{Results}
The upper part of Table~\ref{tab:dexonomy} shows the results of the zero-shot generalization experiments.
\modelname outperforms baseline methods across all settings.
On the challenging \textit{Unseen Both} split where both objects and grasp taxonomy are novel, the model shows reasonable performance degradation.
Note that Dexonomy provides sparse coverage of grasps per instruction compared to DexGYS, as each object-instruction pair admits multiple valid solutions but contains only one ground truth pose, making distance-based metrics like CD less indicative of actual grasp quality.

\input{tables/02_dexonomy}

\subsection{Steerable Grasp Generation}
\label{subsec:steerable}

Beyond standard language-conditioned generation, we propose a steerable grasp generation strategy that provides users with fine-grained control over the grasp configuration.
Our approach leverages the autoregressive nature of \modelname: by providing a partially filled ECoT sequence as context (e.g., specifying 1-2 finger joints and their contact positions), we can guide the model to complete the remaining sequence while respecting these constraints.
Specifically, users can prefix the generation with partial sequences like $\langle$\texttt{thbase}$\rangle \langle p_{x} \rangle \langle p_{y} \rangle \langle p_{z} \rangle \langle$\texttt{ffdistal}$\rangle \langle p_{x} \rangle \langle p_{y} \rangle \langle p_{z} \rangle$, where link tokens identify specific finger links and $\langle p_{*} \rangle$ represents discretized position tokens of links.
The model then completes the remaining contact and action tokens.
This steerable interface enables practical applications where specific contact requirements must be satisfied, such as precision grasps or task-specific manipulations.

\noindentbold{Setup}
We systematically adjust the number of specified links from 1 to 5, and evaluate the intention alignment (P-FID and CD) and grasp quality (Success rate).

\noindentbold{Results}
The bottom part of Table~\ref{tab:dexonomy} shows the results of steerable generation across various partial ECoT constraints.
Providing more steering context leads to generated grasps that more closely align with the constraints, as evidenced by drastically improved intention alignment metrics (lower P-FID and CD).
Notably, the success rate also improves with additional constraints, indicating that steerable generation maintains or even enhances grasp quality while satisfying user-specified contact requirements.
Analyzing generalization patterns across splits, we observe that the model generalizes better to unseen objects than to unseen grasp taxonomies.
In qualitative analysis of steerable generation on unseen grasp taxonomies, we empirically find that generated grasps often hold the object but exhibit instability (e.g., shaking), suggesting that learning stable contact patterns for novel manipulation strategies remains challenging.

\subsection{Additional Analysis}
\label{subsec:additional_analysis}

We provide additional analyses to validate the quality of intermediate contact reasoning, robustness under realistic sensing conditions, and inference speed.
All the experiments are conducted on the DexGYS validation set.

\input{tables/04_ecot_accuracy}
\noindentbold{Contact reasoning quality}
To validate the quality of our embodied contact reasoning, we directly evaluate the predicted contact tokens against ground truth contact annotations.
For contact link prediction, we compare the predicted contact link set against the ground truth using IoU, precision, recall, and F1 scores.
For contact position accuracy, we compare predicted link positions against actual positions computed via forward kinematics, marking a contact as successfully localized if the L2 distance is less than 1cm.
We report Position Accuracy (Pos. Acc.), the percentage of correctly localized contacts across all predictions.

Table~\ref{tab:ecot_accuracy} presents the contact prediction quality on the DexGYS validation set.
The model demonstrates satisfactory performance across both contact link prediction and spatial localization, validating the precision of our contact-based embodied reasoning approach.

\noindentbold{Robustness to partial observations}
We evaluate robustness to partial point cloud observations in Sec.~\ref{subsec:partial_obs}, where \modelname maintains stable performance with only ${\sim}$35\% of points visible and added sensor noise.

\noindentbold{Inference speed}
Table~\ref{tab:inference_time} reports the average inference time on a single NVIDIA A6000 GPU.
\modelname achieves comparable speed to DexGYSNet, with the embodied reasoning step adding only modest overhead.

\input{tables/05_inference_time}

%% file: tables/01_dexgys_eval.tex
\begin{table*}[!t]
    \centering
        \resizebox{0.80\linewidth}{!}{
        \begin{tabular}{l|ccc|ccc|ccc}
            \toprule
            \multirow{2}{*}{Method} & \multicolumn{3}{c|}{Intention} & \multicolumn{3}{c|}{Quality} & \multicolumn{3}{c}{Diversity} \\
            & P-FID$\downarrow$ & CD$\downarrow$ & Con.$\downarrow$ & Success $\uparrow$ & $Q_1$$\uparrow$ & Pen.$\downarrow$ & $\delta_t\uparrow$ & $\delta_r\uparrow$ & $\delta_q\uparrow$ \\
              & & & {\footnotesize($\times10^{-1}$)} & {\footnotesize(\%)} & {\footnotesize($\times10^{-1}$)} & & & & \\
             \midrule
             GraspCVAE~\cite{sohn2015learning} & 29.02 & 3.14 & 0.96 & 29.12 & 0.54 & 0.55 & 0.18 & 1.76 & 0.18 \\
             GraspTTA~\cite{jiang2021hand} & 33.15 & 12.19 & 1.11 & 43.46 & 0.71 & 0.19 & 2.11 & 6.15 & 3.87 \\
             SceneDiffusers~\cite{huang2023diffusion} & 7.93 & 1.68 & 0.45 & 62.24 & 0.83 & 0.25 & 0.35 & 3.46 & 0.39 \\
             DGTR~\cite{xu2024dexterous} & 15.77 & 2.90 & 0.78 & 51.91 & 0.78 & 0.16 & 2.05 & 14.01 & 4.30 \\
             DexGYSNet~\cite{wei2024grasp} & 5.60 & 1.20 & 0.36 & 63.31 & 0.83 & 0.22 & 6.12 & 55.68 & 6.12 \\
            \rowcolor{gray!20}\modelname (w/o ER) & \underline{0.30} & 1.95 & 0.40 & 62.37 & 0.66 & 0.44 & \underline{8.78} & \underline{77.13} & \textbf{13.77} \\
            \rowcolor{gray!20}\modelname & \textbf{0.20} & \underline{1.46} & \textbf{0.34} & \textbf{67.14} & \textbf{0.89} & 0.37 & \textbf{8.84} & \textbf{77.98} & \underline{13.63} \\
            \bottomrule
        \end{tabular}
        }
        \caption{
            \textbf{Language-conditioned grasp generation on DexGYS validation set}. We evaluate intention alignment (P-FID, CD, Con.), physical quality (Success, $Q_1$, Pen.), and diversity ($\delta_t$, $\delta_r$, $\delta_q$). \modelname outperforms all baselines with substantial improvements in intention alignment while maintaining high success rates and diversity. The ablation study (w/o ER) demonstrates that embodied reasoning through contact prediction is critical for aligning grasps with task semantics. %
        }
    \label{tab:dexgys_eval}
\end{table*}

%% file: figures/03_qual/figure.tex
\begin{figure*}[!t]
    \centering
    \includegraphics[width=\linewidth]{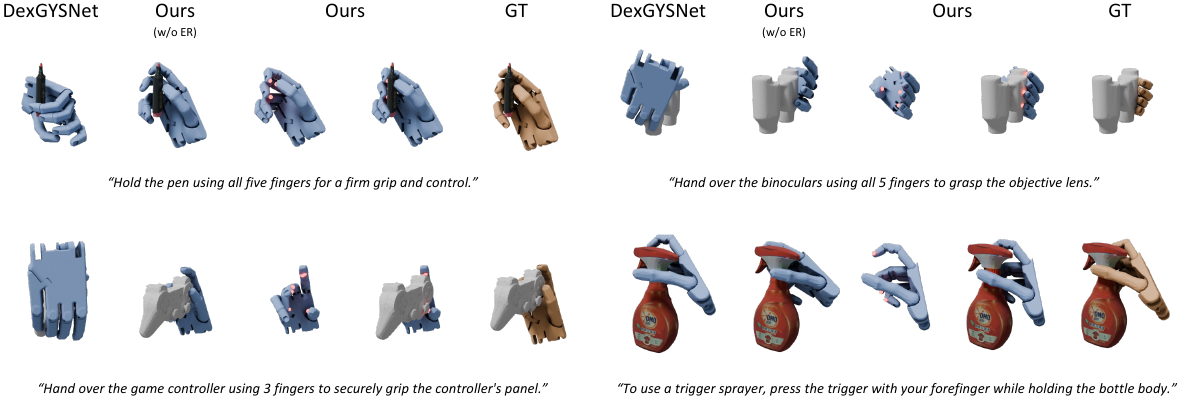}
    \vspace{-7mm}
    \caption{\textbf{Qualitative results on language-conditioned dexterous grasp generation.}
    Given object point clouds and natural language instructions, \modelname generates embodied contact predictions (shown as colored spheres on object surfaces) followed by grasp configurations.
    The model successfully captures task-specific contact patterns and produces physically plausible grasps that align with language instructions across diverse objects and manipulation intents.
    The 3rd and 4th columns show predictions from the same model (\modelname), visualized in two separate columns to better highlight the predicted contact points in the 3rd column.}
    \label{fig:qual}
\end{figure*}

%% file: tables/03_ablation.tex
\begin{table}[t]
    \centering
    \resizebox{\linewidth}{!}{
    \begin{tabular}{l|ccc|c}
        \toprule
        Configuration & P-FID$\downarrow$ & CD$\downarrow$ & Con.$\downarrow$ & Success$\uparrow$ \\
         & & & {\footnotesize($\times10^{-1}$)} & {\footnotesize(\%)} \\
        \midrule
        \rowcolor{gray!20}\multicolumn{5}{l}{\textit{ECoT}} \\
        \quad w/o ECoT & 0.30 & 1.95 & 0.40 & 62.37 \\
        \quad w/ ECoT (default) & \textbf{0.20} & \textbf{1.46} & \textbf{0.34} & \textbf{67.14} \\
        \midrule
        \rowcolor{gray!20}\multicolumn{5}{l}{\textit{Action bin sizes ($N_\mathbf{a}$)}} \\
        \quad $N_\mathbf{a}=128$ & 0.21 & 1.70 & 0.35 & 66.19 \\
        \quad $N_\mathbf{a}=256$ (default) & \textbf{0.20} & \textbf{1.46} & \textbf{0.34} & \textbf{67.14} \\
        \quad $N_\mathbf{a}=512$ & 0.26 & 1.89 & 0.37 & 65.24\\
        \midrule
        \rowcolor{gray!20}\multicolumn{5}{l}{\textit{Position bin sizes ($N_{\text{pos}}$)}} \\
        \quad $N_{\text{pos}}=128$ & 0.21 & 1.67 & 0.36 & \textbf{67.35}\\
        \quad $N_{\text{pos}}=256$ (default) & \textbf{0.20} & \textbf{1.46} & \textbf{0.34} & 67.14 \\
        \quad $N_{\text{pos}}=512$ & 0.23 & 1.69 & 0.35 & 66.32\\
        \midrule
        \rowcolor{gray!20}\multicolumn{5}{l}{\textit{Contact position dropout}} \\
        \quad $p_{\text{drop}}=0.0$ (no dropout) & 0.22 & 1.67 & 0.36 & 65.68\\
        \quad $p_{\text{drop}}=0.5$ (default) & \textbf{0.20} & \textbf{1.46} & \textbf{0.34} & \textbf{67.14} \\
        \quad $p_{\text{drop}}=0.7$ & 0.25 & 1.85 & 0.37 & 66.54\\
        \quad $p_{\text{drop}}=1.0$ & 0.30 & 2.10 & 0.40 & 63.33\\
        \midrule
        \rowcolor{gray!20}\multicolumn{5}{l}{\textit{Point cloud encoder}} \\
        \quad Uni3D~\cite{zhou2024uni3d} & 0.52 & 2.24 & 0.46 & 59.07\\
        \quad PartField~\cite{liu2025partfield} (default) & \textbf{0.20} & \textbf{1.46} & \textbf{0.34} & \textbf{67.14} \\
        \midrule
        \rowcolor{gray!20}\multicolumn{5}{l}{\textit{LLM backbone}} \\
        \quad Qwen2.5-0.5B~\cite{qwen2.5} (default) & \textbf{0.20} & \textbf{1.46} & \textbf{0.34} & 67.14 \\
        \quad Qwen2.5-1.5B~\cite{qwen2.5} & 0.18 & 1.52 & -- & \textbf{67.55} \\
        \quad SmolLM2-360M~\cite{allal2025smollm2} & 0.31 & 2.24 & -- & 64.87 \\
        \bottomrule
    \end{tabular}
    }
    \caption{
        \textbf{Ablation studies on DexGYS validation set}. We systematically analyze the impact of key design choices: token discretization granularity, point cloud encoder architecture, backbone model capacity, and contact position dropout probability. We report primary metrics for intention alignment (P-FID, Con.) and physical quality (Success). The default configuration achieves strong performance across all metrics.
    }
    \label{tab:ablation}
\end{table}

%% file: tables/02_dexonomy.tex
\begin{table*}[!t]
    \centering
        \resizebox{\linewidth}{!}{
        \begin{tabular}{l|ccc|ccc|ccc|ccc}
            \toprule
            & \multicolumn{3}{c|}{Seen Obj.\& Grasp} & \multicolumn{3}{c|}{Unseen Obj.} & \multicolumn{3}{c|}{Unseen Grasp} & \multicolumn{3}{c}{Unseen Obj.\& Grasp} \\
            & P-FID$\downarrow$ & CD$\downarrow$ & Success (\%)$\uparrow$ & P-FID$\downarrow$ & CD$\downarrow$ & Success (\%)$\uparrow$ & P-FID$\downarrow$ & CD$\downarrow$ & Success (\%)$\uparrow$ & P-FID$\downarrow$ & CD$\downarrow$ & Success (\%)$\uparrow$ \\
            \midrule
            DexGYS~\cite{wei2024grasp} & 1.89 & 19.45 & 0.97 & 2.55 & 20.03 & 0.93 & 2.87 & 20.39 & 0.99 & 3.04 & 20.89 & 1.16 \\
            DextER & 0.44 & 18.32 & 12.24 & 1.44 & 20.07 & 10.86 & 1.04 & 19.49 & 9.10 & 1.23 & 20.44 & 8.41 \\
            $\vdash$ \textit{1 link} & 0.43 & 5.51 & 10.40 & 0.41 & 5.56 & 10.19 & 1.06 & 5.93 & 7.08 & 1.03 & 6.09 &7.23 \\
            $\vdash$ \textit{2 links} & 0.28 & 2.33 & 14.67 & 0.27 & 2.50 & 14.36 & 1.02 & 3.78 & 7.50 & 0.99 & 3.68 &7.55 \\ 
            $\vdash$ \textit{3 links} & 0.18 & 1.50 & 17.84 & 0.19 & 1.57 & 17.32 & 0.93 & 2.57 & 8.33 & 0.87 & 2.55 &8.36 \\    
            $\vdash$ \textit{4 links} & 0.14 & 0.91 & 20.14 & 0.14 & 1.04 & 19.33 & 0.85 & 2.19 & 8.38 & 0.83 & 2.17 & 8.82 \\ 
            $\vdash$ \textit{5 links} &  0.12 & 0.73 &21.35 & 0.14 & 0.88 & 20.26 & 0.83 & 1.93 & 9.10 & 0.78 & 1.98 & 8.31 \\ 
            \bottomrule
        \end{tabular}
        }
        \caption{
            \textbf{Evaluation on Dexonomy~\cite{chen2025dexonomy}.}
            (Upper) Zero-shot generalization across data splits.
            (Lower) Steerable generation with partial contact specification.
        }
    \label{tab:dexonomy}
\end{table*}

%% file: tables/04_ecot_accuracy.tex
\begin{table}[t]
    \centering
    \resizebox{0.9\linewidth}{!}{
    \begin{tabular}{l|ccccc}
        \toprule
        Method & IoU$\uparrow$ & Precision$\uparrow$ & Recall$\uparrow$ & F1$\uparrow$ & Pos. Acc.$\uparrow$ \\
        \midrule
        \modelname & 0.42 & 0.59 & 0.63 & 0.57 & 0.79 \\
        \bottomrule
    \end{tabular}
    }
    \caption{
        \textbf{Contact prediction quality}. We evaluate the accuracy of predicted contact links and positions in the embodied CoT sequence. IoU, Precision, Recall, and F1 measure the contact link prediction accuracy. Position Accuracy measures the percentage of predicted contact positions within 1cm of the actual link positions computed via forward kinematics of the predicted grasp pose.
    }
    \label{tab:ecot_accuracy}
\end{table}

%% file: tables/05_inference_time.tex
\begin{table}[h]
\centering
\resizebox{0.90\linewidth}{!}{
\begin{tabular}{@{}lccc@{}}
\toprule
& DexGYSNet & DextER w/o ER & DextER (Ours) \\
\midrule
Time (s) & 1.134 & 1.105 & 1.458 \\
\bottomrule
\end{tabular}
}
\caption{Inference time comparison on DexGYS.}
\label{tab:inference_time}
\end{table}

%% file: sec/5_conclusion.tex
\section{Conclusion}
\label{sec:conclusion}

We presented \modelname, a vision-language approach for language-conditioned dexterous grasp generation that leverages embodied reasoning through contact prediction.
Our method achieves 67.14\% grasp success rate on DexGYS, outperforming prior state-of-the-art by 3.83 p.p., with a 96.4\% improvement in intention alignment, demonstrating that contact reasoning is critical for understanding task semantics and producing diverse, stable grasp configurations.
The autoregressive generation framework enables steerable grasp generation, where users can guide the model by specifying partial contact constraints, providing an effective interface for fine-grained control over grasp synthesis.

\noindentbold{Limitations}
Our autoregressive framework is susceptible to compounding errors, which diffusion-based approaches might mitigate, though combining these with embodied reasoning remains underexplored.
Current evaluation focuses on simulation with single, static objects, though zero-shot robustness to partial and noisy point clouds is encouraging for real-world deployment; extending to complex scenes with occlusions would better reflect real-world scenarios.
Sequential token prediction may also limit real-time performance.

%% file: sec/X_suppl.tex
\clearpage
\setcounter{page}{1}
\maketitlesupplementary

\section{Implementation Details}
\label{sec:impl_details}
We provide additional implementation details of \modelname.

\subsection{Model}
\label{subsec:model_arch}

\noindentbold{Point cloud encoder}
We support two pretrained 3D point cloud encoders in our experiments:
\begin{itemize}[leftmargin=*, itemsep=1pt, parsep=0pt, topsep=0pt]
    \item \textbf{PartField}~\cite{liu2025partfield}: A triplane-based encoder pretrained on Objaverse using contrastive learning with SAM2 masks for part-aware feature extraction. The encoder produces part-geometry aware features beneficial for contact reasoning. We downsample the triplane feature maps at the encoder bottleneck, yielding 768 visual tokens (3 $\times$ 16 $\times$ 16).
    \item \textbf{Uni3D}~\cite{zhou2024uni3d}: A transformer-based point cloud encoder pretrained on OpenShape~\cite{liu2023openshape} dataset. This encoder provides global object-level features, where we use FPS sampling with 512 points processed through pretrained CLIP vision transformer layers to produce visual tokens. We use the `base' variant of the encoder.
\end{itemize}
Both encoders process input point clouds sampled to 10,000 points with XYZ coordinates.

\vspace{0.3cm}
\noindentbold{Tokenization}
We discretize continuous values into tokens:
\begin{itemize}[leftmargin=*, itemsep=1pt, parsep=0pt, topsep=0pt]
    \item \textbf{Action tokens}: Grasp actions are normalized using quantile normalization (1st-99th percentile $\rightarrow$ $[-1, 1]$), then uniformly discretized into $N_{\mathbf{a}}=256$ bins per dimension.
    \item \textbf{Position tokens}: Contact positions are bounded by dataset-specific min/max values computed from the training set, then uniformly discretized into $N_{\text{pos}}=256$ bins per dimension.
    \item \textbf{Link tokens}: Shadow Hand link names are encoded as special tokens (e.g., \texttt{<rh\_ffdistal>}, \texttt{<rh\_palm>}, \texttt{<rh\_thdistal>}).
    \item \textbf{Delimiter tokens}: Special tokens mark sequence boundaries: \texttt{<|contact\_start|>}, \texttt{<|contact\_end|>}, \texttt{<|action\_start|>}, \texttt{<|action\_end|>}.
\end{itemize}
All special tokens are registered in the pretrained tokenizer, expanding the vocabulary while preserving language understanding capabilities.

\vspace{0.3cm}
\noindentbold{Attention mechanism}
We employ a hybrid attention pattern where point cloud tokens use bidirectional attention to capture global geometric context, while the other tokens follow causal (autoregressive) attention. This design enables comprehensive 3D understanding while maintaining standard next-token prediction for text and actions. The attention mask pattern is illustrated in Fig.~\ref{fig:attention_mask}.

\input{figures/04_attention_mask/figure}

\subsection{Training}
\label{subsec:training_details}

We initialize the model from pretrained point cloud encoder and LLM backbone checkpoints, then finetune all components end-to-end.
The training hyperparameters are summarized in Table~\ref{tab:training_config}.
\input{tables/06_training_detail}

\noindentbold{Prompt design}
We use ChatML-style conversation prompt consistent with the Qwen2.5 model, the base LLM model \modelname uses. Table~\ref{tab:prompt_example} shows an example of the prompt.
\input{tables/07_prompt}

\subsection{Dataset}
\label{subsec:dataset_curation}
\noindentbold{Contact annotation}
For each grasp in the DexGYS and Dexonomy datasets, we automatically extract contact annotations using MuJoCo physics simulation. We load the Shadow Hand and object models into MuJoCo, execute forward kinematics for each grasp pose, and use precise collision queries between finger links and object meshes to extract the 3D surface positions where each hand link makes contact with the object. Since these annotations are obtained directly from the physics engine, they are physically grounded and accurate to simulator precision. These contact annotations enable our model to reason about grasp contact patterns during training. Fig.~\ref{fig:contact_annotation} shows an example of this annotation.
\input{figures/05_contact_annotation/figure}

\noindentbold{Grasp instruction annotation}
To annotate Dexonomy with grasp instructions, we provide Gemma-3~\cite{team2025gemma} with multi-view renderings of each hand–object configuration with the annotated contact link set. Conditioned on this input and the structured prompt template in Fig.~\ref{fig:gemma_prompt_template}, the model first infers the object’s semantic category and identifies the functional part being contacted. It then generates two complementary descriptions: (1) a low-level physical grasp instruction and (2) a high-level functional instruction. For all Dexonomy experiments, we use only the physical grasp instructions. This multi-cue conditioning (visual evidence, contact set, and chain-of-thought prompting) serves as our annotation validity mechanism, as the model must produce descriptions consistent across all input modalities. Fig.~\ref{fig:grasp_instruction_annotation_example} presents an example of our grasp instruction annotation.

We note that our primary design choice for Dexonomy annotations is to generate open-vocabulary descriptions that capture fine-grained finger-object contact patterns from visual observations, rather than descriptions limited to a fixed grasp taxonomy.
While incorporating taxonomy labels could enhance semantic expressiveness, our image-conditioned VLM approach yields more detailed, contact-aware descriptions that enable precise control in downstream tasks.
\subsection{Evaluation}
\label{subsec:eval_metrics}

\noindentbold{Evaluation metrics}
Following \cite{wei2024grasp}, we evaluate generated grasps using the following metrics:
\begin{itemize}[leftmargin=*, itemsep=1pt, parsep=0pt, topsep=0pt]
    \item \textbf{P-FID}: Fréchet Distance between point cloud features~\cite{nichol2022point} of generated and reference grasps, measuring distributional similarity.
    \item \textbf{Chamfer Distance (CD)}: Average spatial discrepancy between generated and ground-truth hand meshes.
    \item \textbf{Contact Distance (Con.)}: L2 distance between predicted and target contact maps on the object surface.
    \item \textbf{Success Rate}: Percentage of grasps that remain stable in Isaac Gym simulation after execution.
    \item $\mathbf{Q}_1$: Force-closure quality metric measuring the minimum wrench required to break grasp stability.
    \item \textbf{Penetration (Pen.)}: Maximum penetration depth between hand mesh and object point cloud.
    \item \textbf{Diversity} ($\delta_t$, $\delta_r$, $\delta_q$): Standard deviation of palm translation, rotation, and joint angles across generated samples.
\end{itemize}
P-FID, CD, and Con. measure how well grasps align with task intentions, while Success Rate, $Q_1$, and Pen. assess physical plausibility and stability.
Diversity metrics evaluate the model's ability to generate varied grasp configurations.

\noindentbold{Success rate criteria}
We evaluate grasp success using different criteria depending on the benchmark:
\begin{itemize}[leftmargin=*, itemsep=1pt, parsep=0pt, topsep=0pt]
    \item \textbf{DexGraspNet}~\cite{wei2024grasp}: A grasp succeeds if the hand maintains stable contact with the object for 100 simulation steps under at least one of six gravity directions, with maximum penetration depth $\leq$ 0.1 cm.
    \item \textbf{DexGraspBench}~\cite{chen2025bodex}: A grasp succeeds if it resists six external forces in MuJoCo simulation without severe penetrations (max 1 cm), and the object's final pose remains within 5 cm translation and 15° rotation from its initial configuration.
\end{itemize}

\vspace{0.3cm}
\noindentbold{Dexonomy data splits}
To analyze both taxonomy-level and object-level generalization, we structure the Dexonomy dataset into five splits with the following sample counts:

\begin{itemize}[leftmargin=*, itemsep=1pt, parsep=0pt, topsep=0pt]
\item \textbf{Training}: 182,061
\item \textbf{Seen Objects \& Taxonomy}: 12,929
\item \textbf{Unseen Objects }: 12,929
\item \textbf{Unseen Grasp Taxonomy}: 12,929
\item \textbf{Unseen Both}: 12,929
\end{itemize}
The definitions of seen and unseen grasp taxonomies and objects are as follows:

\begin{itemize}[leftmargin=*, itemsep=1pt, parsep=0pt, topsep=0pt]
\item \textbf{Seen grasp taxonomies}: Small Diameter, Medium Wrap, Power Disk, Power Sphere, Precision Disk, Precision Sphere, Tip Pinch, Light Tool, Writing Tripod, Parallel Extension, Lateral Tripod, Quadpod, Stick, Prismatic 4 Finger, Prismatic 3 Finger, Tripod, Fixed Hook, Lateral, Index Finger Extension, Extensior Type, Palmar, Ring, Ventral, Inferior Pincer
\item \textbf{Unseen grasp taxonomies}: Large Diameter, Adducted Thumb, Prismatic 2 Finger, Adduction Grip, Sphere 4 Finger, Sphere 3 Finger
\item \textbf{Seen objects}: Objects included in the training portion of the Dexonomy dataset.
\item \textbf{Unseen objects}: Objects included in the test portion of the Dexonomy dataset.
\end{itemize}

\input{figures/08_grasp_annotation/figure}

\subsection{Robustness to Partial Observations}
\label{subsec:partial_obs}

To evaluate robustness under realistic sensing conditions, we simulate partial point cloud observations using the Hidden Point Removal (HPR) algorithm~\cite{katz2007direct}, which determines point visibility from a given viewpoint.
We place two virtual cameras following common VLA hardware setups: (1) an \textit{egocentric} wrist-mounted camera positioned behind the hand looking toward the object, and (2) a \textit{third-person} camera providing an overhead/side view.
This dual-camera setup retains only ${\sim}$35\% of the original points visible from these viewpoints.
We further add depth-dependent Gaussian noise ($\sigma_d{=}$4mm, scaling linearly with distance) and lateral noise ($\sigma_l{=}$1mm) to simulate real sensor artifacts.

\input{figures/05_rebuttal/figure}

This evaluation is conducted in a zero-shot setting without retraining on partial or noisy inputs.
As shown in Table~\ref{tab:partial_obs}, performance degrades only slightly (Success rate $-$2\%, P-FID $+$5\%, CD $+$3\%) despite the severe reduction in observed points and added noise, demonstrating that \modelname generalizes robustly beyond perfect point cloud observations.

\begin{table}[h]
\small
\centering
\begin{tabular}{@{}lcc@{}}
\toprule
& Full & Partial + Noisy \\
\midrule
Success$\uparrow$ (\%) & 67.14 & 65.77 \\
P-FID$\downarrow$ & 0.20 & 0.21 \\
CD$\downarrow$ & 1.46 & 1.50 \\
\bottomrule
\end{tabular}
\caption{
    \textbf{Robustness to partial and noisy observations on DexGYS.} Zero-shot evaluation without retraining on partial inputs.
}
\label{tab:partial_obs}
\end{table}

\section{Qualitative Analysis}
\label{sec:qualitative_analysis}

We provide additional qualitative analysis of \modelname on the DexGYS (Fig.~\ref{fig:suppl_qual}) and Dexonomy datasets.(Fig.~\ref{fig:suppl_qual_dexonomy})
We further present qualitative results for steerable grasp generation. As illustrated in Fig.~\ref{fig:steerable_grasp_generation}, increasing the number of ECoT constraints from Steer-1 to Steer-5 progressively guides the model toward producing grasps that more closely resemble the ground-truth configuration.

\subsection{Failure Mode Analysis}
\label{subsec:failure_analysis}

We identify two consistent failure patterns in \modelname:

\noindentbold{Penetration-induced failures}
Quantization artifacts from discretizing continuous contact and grasp parameters into token bins account for 14.7\% of failure cases.
In these cases, the predicted grasp is semantically correct---the model correctly identifies which fingers should contact which object regions---but the resulting hand configuration exhibits slight penetration with the object mesh, causing simulation failure.
We intentionally avoid geometric post-processing (\eg, optimization-based refinement commonly used in baseline methods) to analyze the pure effect of autoregressive contact reasoning.

\noindentbold{Unstable grasps on unseen taxonomy}
In unseen grasp taxonomy experiments, \modelname predicts plausible poses (reflected in low Chamfer Distance and P-FID), but physical execution becomes unstable (\eg, object shaking during simulation).
This suggests that while the model generalizes contact reasoning to novel grasp types, subtle structure-dependent stability cues---such as force closure properties specific to certain grasp taxonomies---are not fully captured by the training supervision.

\subsection{Steerable Generation: Practical Use Cases}
\label{subsec:steerable_usecases}

Our contact-based steering interface offers fine-grained, unambiguous control over grasp synthesis.
Given full point cloud observations, sampling contact points on the object surface is straightforward using standard processing (\eg, farthest point sampling, region selection).
This interface provides significantly more precise control than implicit text descriptions and is valuable in several practical scenarios:
(1) collecting diverse manipulation data by varying contact specifications,
(2) task-conditioned grasping (\eg, gripping a tool at a functional position),
(3) planning grasps in constrained environments where specific contact regions must be avoided, and
(4) debugging grasp failures by systematically varying contact constraints.
\input{figures/10_suppl_qual/figure}
\input{figures/06_dexonomy_qual/figure}
\input{figures/09_steerable_grasp/figure}

\input{figures/07_gemma_template/figure}

%% file: figures/04_attention_mask/figure.tex
\begin{figure}[t]
    \centering
    \begin{tikzpicture}[scale=0.7]
        \definecolor{attendcolor}{RGB}{70, 130, 180}  %
        \definecolor{maskcolor}{RGB}{220, 220, 220}   %

        \def\bossize{3}   %
        \def\pcsize{3}    %
        \def\textsize{3}  %
        \def\contsize{3}  %
        \def\actsize{3}   %
        \def\totalsize{15} %
        \def\cellsize{0.5} %

        \pgfmathsetmacro{\gridsize}{\totalsize*\cellsize}
        \pgfmathsetmacro{\boswidth}{\bossize*\cellsize}
        \pgfmathsetmacro{\pcwidth}{\pcsize*\cellsize}
        \pgfmathsetmacro{\textwidth}{\textsize*\cellsize}
        \pgfmathsetmacro{\contwidth}{\contsize*\cellsize}
        \pgfmathsetmacro{\actwidth}{\actsize*\cellsize}
        \pgfmathsetmacro{\bosstart}{\gridsize-\boswidth}
        \pgfmathsetmacro{\pcstart}{\bosstart-\pcwidth}
        \pgfmathsetmacro{\textstart}{\pcstart-\textwidth}
        \pgfmathsetmacro{\contstart}{\textstart-\contwidth}
        \pgfmathsetmacro{\actstart}{\contstart-\actwidth}
        \draw[step=\cellsize, gray, very thin] (0,0) grid (\gridsize, \gridsize);

        \foreach \i in {0,...,2} {
            \pgfmathsetmacro{\yend}{\bosstart+(3-\i)*\cellsize}
            \pgfmathsetmacro{\xend}{(\i+1)*\cellsize}
            \fill[attendcolor] (0, \bosstart) rectangle (\xend, \yend);
        }

        \fill[attendcolor] (0,\pcstart) rectangle (\boswidth, \bosstart);
        \fill[attendcolor] (\boswidth,\pcstart) rectangle (\boswidth+\pcwidth, \bosstart);

        \fill[attendcolor] (0,\textstart) rectangle (\boswidth, \pcstart);
        \fill[attendcolor] (\boswidth,\textstart) rectangle (\boswidth+\pcwidth, \pcstart);
        \foreach \i in {0,...,2} {
            \pgfmathsetmacro{\yend}{\textstart+(3-\i)*\cellsize}
            \pgfmathsetmacro{\xend}{\boswidth+\pcwidth+(\i+1)*\cellsize}
            \fill[attendcolor] (\boswidth+\pcwidth, \textstart) rectangle (\xend, \yend);
        }

        \fill[attendcolor] (0,\contstart) rectangle (\boswidth, \textstart);
        \fill[attendcolor] (\boswidth,\contstart) rectangle (\boswidth+\pcwidth, \textstart);
        \fill[attendcolor] (\boswidth+\pcwidth,\contstart) rectangle (\boswidth+\pcwidth+\textwidth, \textstart);
        \foreach \i in {0,...,2} {
            \pgfmathsetmacro{\yend}{\contstart+(3-\i)*\cellsize}
            \pgfmathsetmacro{\xend}{\boswidth+\pcwidth+\textwidth+(\i+1)*\cellsize}
            \fill[attendcolor] (\boswidth+\pcwidth+\textwidth, \contstart) rectangle (\xend, \yend);
        }

        \fill[attendcolor] (0,0) rectangle (\boswidth, \contstart);
        \fill[attendcolor] (\boswidth,0) rectangle (\boswidth+\pcwidth, \contstart);
        \fill[attendcolor] (\boswidth+\pcwidth,0) rectangle (\boswidth+\pcwidth+\textwidth, \contstart);
        \fill[attendcolor] (\boswidth+\pcwidth+\textwidth,0) rectangle (\boswidth+\pcwidth+\textwidth+\contwidth, \contstart);
        \foreach \i in {0,...,2} {
            \pgfmathsetmacro{\yend}{\actstart+(3-\i)*\cellsize}
            \pgfmathsetmacro{\xend}{\boswidth+\pcwidth+\textwidth+\contwidth+(\i+1)*\cellsize}
            \fill[attendcolor] (\boswidth+\pcwidth+\textwidth+\contwidth, \actstart) rectangle (\xend, \yend);
        }

        \draw[step=\cellsize, black, very thin] (0,0) grid (\gridsize, \gridsize);

        \draw[black, thick] (\boswidth, 0) -- (\boswidth, \gridsize);
        \draw[black, thick] (\boswidth+\pcwidth, 0) -- (\boswidth+\pcwidth, \gridsize);
        \draw[black, thick] (\boswidth+\pcwidth+\textwidth, 0) -- (\boswidth+\pcwidth+\textwidth, \gridsize);
        \draw[black, thick] (\boswidth+\pcwidth+\textwidth+\contwidth, 0) -- (\boswidth+\pcwidth+\textwidth+\contwidth, \gridsize);
        \draw[black, thick] (0, \bosstart) -- (\gridsize, \bosstart);
        \draw[black, thick] (0, \pcstart) -- (\gridsize, \pcstart);
        \draw[black, thick] (0, \textstart) -- (\gridsize, \textstart);
        \draw[black, thick] (0, \contstart) -- (\gridsize, \contstart);

        \draw[black, ultra thick] (0,0) rectangle (\gridsize, \gridsize);

        \pgfmathsetmacro{\boslabelx}{\boswidth/2}
        \pgfmathsetmacro{\pclabelx}{\boswidth+\pcwidth/2}
        \pgfmathsetmacro{\textlabelx}{\boswidth+\pcwidth+\textwidth/2}
        \pgfmathsetmacro{\contlabelx}{\boswidth+\pcwidth+\textwidth+\contwidth/2}
        \pgfmathsetmacro{\actlabelx}{\boswidth+\pcwidth+\textwidth+\contwidth+\actwidth/2}
        \node at (\boslabelx, \gridsize+0.4) {\small \textbf{$\text{Text}_\text{pre}$}};
        \node at (\pclabelx, \gridsize+0.4) {\small \textbf{PC}};
        \node at (\textlabelx, \gridsize+0.4) {\small \textbf{$\text{Text}_\text{post}$}};
        \node at (\contlabelx, \gridsize+0.4) {\small \textbf{Cont}};
        \node at (\actlabelx, \gridsize+0.4) {\small \textbf{Act}};

        \pgfmathsetmacro{\boslabely}{\bosstart+\boswidth/2}
        \pgfmathsetmacro{\pclabely}{\pcstart+\pcwidth/2}
        \pgfmathsetmacro{\textlabely}{\textstart+\textwidth/2}
        \pgfmathsetmacro{\contlabely}{\contstart+\contwidth/2}
        \pgfmathsetmacro{\actlabely}{\actwidth/2}
        \node[rotate=90] at (-0.4, \boslabely) {\small \textbf{$\text{Text}_\text{pre}$}};
        \node[rotate=90] at (-0.4, \pclabely) {\small \textbf{PC}};
        \node[rotate=90] at (-0.4, \textlabely) {\small \textbf{$\text{Text}_\text{post}$}};
        \node[rotate=90] at (-0.4, \contlabely) {\small \textbf{Cont}};
        \node[rotate=90] at (-0.4, \actlabely) {\small \textbf{Act}};

        \pgfmathsetmacro{\keylabelx}{\gridsize/2}
        \pgfmathsetmacro{\querylabely}{\gridsize/2}
        \node at (\keylabelx, \gridsize+0.8) {\textbf{Key}};
        \node[rotate=90] at (-0.8, \querylabely) {\textbf{Query}};

        \pgfmathsetmacro{\legendx}{\gridsize+0.5}
        \pgfmathsetmacro{\legendy}{\gridsize-0.5}
        \begin{scope}[shift={(\legendx, \legendy)}]
            \fill[attendcolor] (0,0) rectangle (0.4,0.4);
            \node[anchor=west] at (0.5, 0.2) {\small Can attend};

            \fill[maskcolor, opacity=0.5] (0,-0.8) rectangle (0.4,-0.4);
            \node[anchor=west] at (0.5, -0.6) {\small Masked};
        \end{scope}

    \end{tikzpicture}
    \caption{\textbf{Prefix-LM attention mask for \modelname.} Point cloud (PC) tokens use bidirectional attention (full blue blocks in PC rows/columns), whereas the other tokens use causal attention (lower triangular patterns), attending to all preceding point cloud tokens.}
    \label{fig:attention_mask}
\end{figure}

%% file: tables/06_training_detail.tex
\begin{table}[h]
    \centering
    \small
    \begin{tabular}{ll}
    \toprule
    \textbf{Parameter} & \textbf{Value} \\
    \midrule
    \rowcolor{gray!10}\multicolumn{2}{l}{\textit{Optimization}} \\
    Learning rate & $1 \times 10^{-4}$ with cosine decay \\
    Warmup steps & 1,000 (linear warmup) \\
    Weight decay & $1 \times 10^{-10}$ \\
    Gradient clipping & Max norm 1.0 \\
    Beta parameters & $\beta_1=0.9$, $\beta_2=0.95$ \\
    Epsilon & $1 \times 10^{-8}$ \\
    \midrule
    \rowcolor{gray!10}\multicolumn{2}{l}{\textit{Training configuration}} \\
    Batch size & 64 (8 per GPU $\times$ 8 GPUs) \\
    Training steps & 100,000 iterations \\
    Precision & Mixed-precision (bfloat16) \\
    Gradient checkpointing & Enabled \\
    Hardware & 8 $\times$ NVIDIA A6000 (48GB) \\
    Framework & PyTorch 2.7+ with HuggingFace \\
    \bottomrule
    \end{tabular}
    \caption{Training hyperparameters and configuration for \modelname.}
    \label{tab:training_config}
\end{table}

%% file: tables/07_prompt.tex
\begin{table*}[!h]
    \centering
    \resizebox{\linewidth}{!}{
    \begin{tabular}{r l  }\toprule
      \bf  Input & \prompttoken{blue!8}{<|im\_start|>}system \ntoken You are Qwen, created by Alibaba Cloud. You are a helpful assistant.\prompttoken{blue!8}{<|im\_end|>} \ntoken \\
                 & \prompttoken{blue!8}{<|im\_start|>}user \ntoken \prompttoken{green!15}{<|vision\_start|>} \prompttoken{green!15}{<|vision\_pad|>} \prompttoken{green!15}{<|vision\_end|>} Think about which hand joints will touch the object and where, then plan the grasp. \\
                 & Query: Hand over the cylinder bottle using all five fingers to grasp the body securely.\prompttoken{blue!8}{<|im\_end|>} \ntoken \\
                 \midrule
    \bf Output  & \prompttoken{blue!8}{<|im\_start|>}assistant \ntoken \jointtoken{<|contact\_start|>} \prompttoken{orange!30}{<|rh\_ffdistal|>}\postoken{<|pos\_bin\_124|>} \postoken{<|pos\_bin\_112|>} \postoken{<|pos\_bin\_129|>} $\cdots$ \jointtoken{<|contact\_end|>} \\
                & \actiontoken{<|action\_start|>} \postoken{<|action\_bin\_171|>} \postoken{<|action\_bin\_148|>} \postoken{<|action\_bin\_61|>} $\cdots$ \actiontoken{<|action\_end|>} \prompttoken{blue!8}{<|im\_end|>} \ntoken \\
                \bottomrule
    \end{tabular}
    }
    \caption{\textbf{Prompt example for \modelname.} Input includes system prompt, vision tokens, and task query. Output contains predicted contacts and action sequences. The model is trained with special delimiter tokens for contact reasoning and actions, where all continuous link positions and actions are normalized as described in \Sref{subsec:impl_details}.}
    \label{tab:prompt_example}
\end{table*}

%% file: figures/05_contact_annotation/figure.tex
\begin{figure}[!t]
    \centering
    \includegraphics[width=\linewidth]{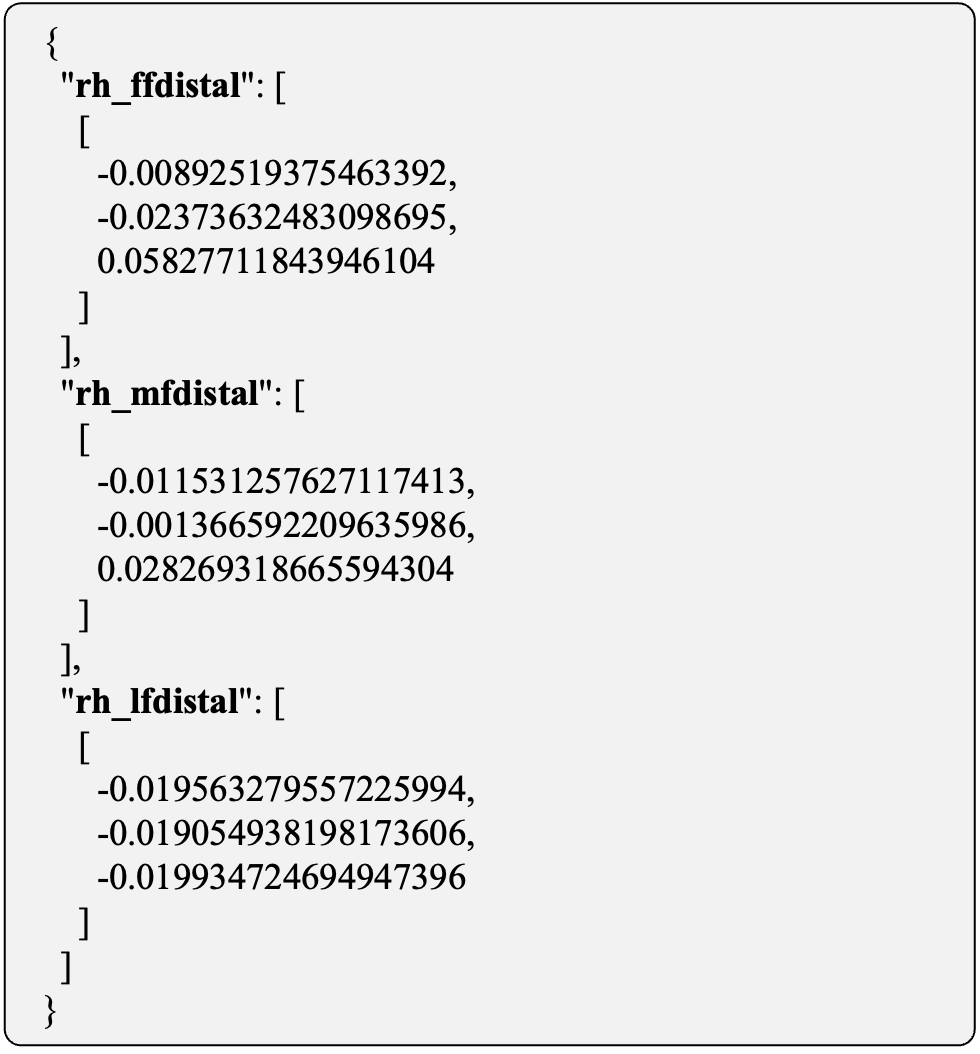}
    \caption{\textbf{Contact annotation example.}
    }
    \label{fig:contact_annotation}
\end{figure}

%% file: figures/08_grasp_annotation/figure.tex
\begin{figure}[ht!]
    \centering
    \includegraphics[width=\linewidth]{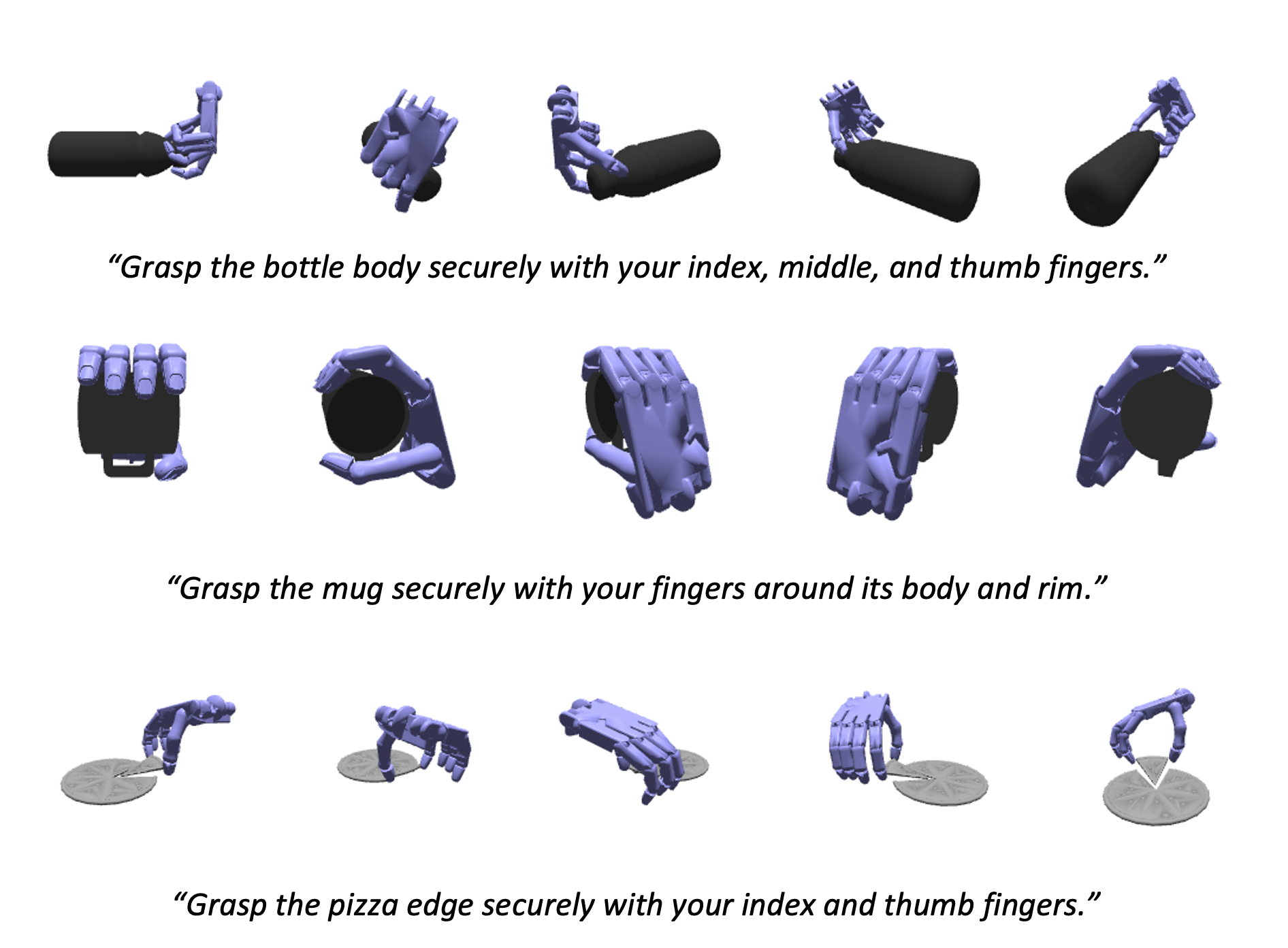}
    \caption{Grasp instruction annotation for the Dexonomy dataset.}
    \label{fig:grasp_instruction_annotation_example}
\end{figure}

%% file: figures/05_rebuttal/figure.tex
\begin{figure}[t]
    \centering
    \includegraphics[width=0.3\linewidth]{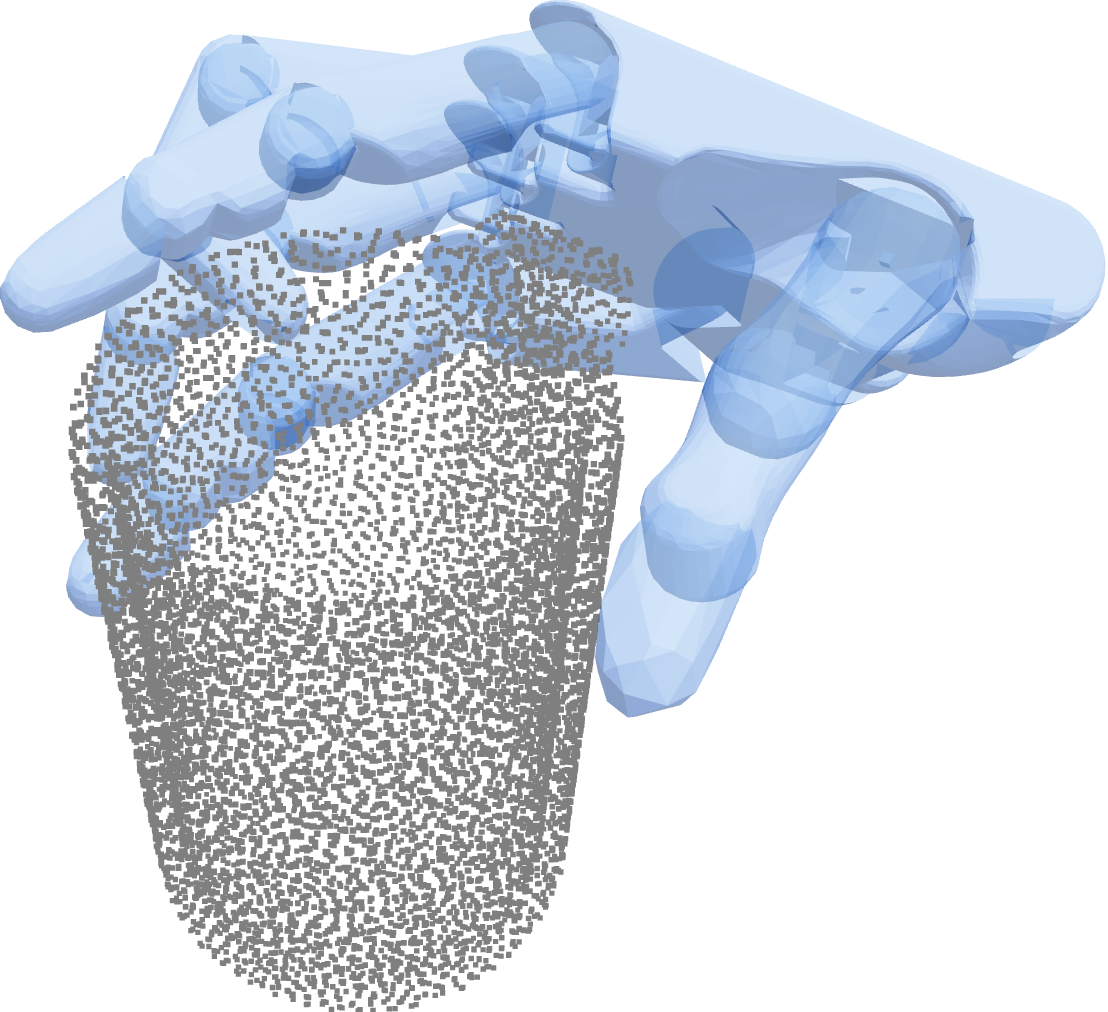}%
    \hfill
    \includegraphics[width=0.3\linewidth]{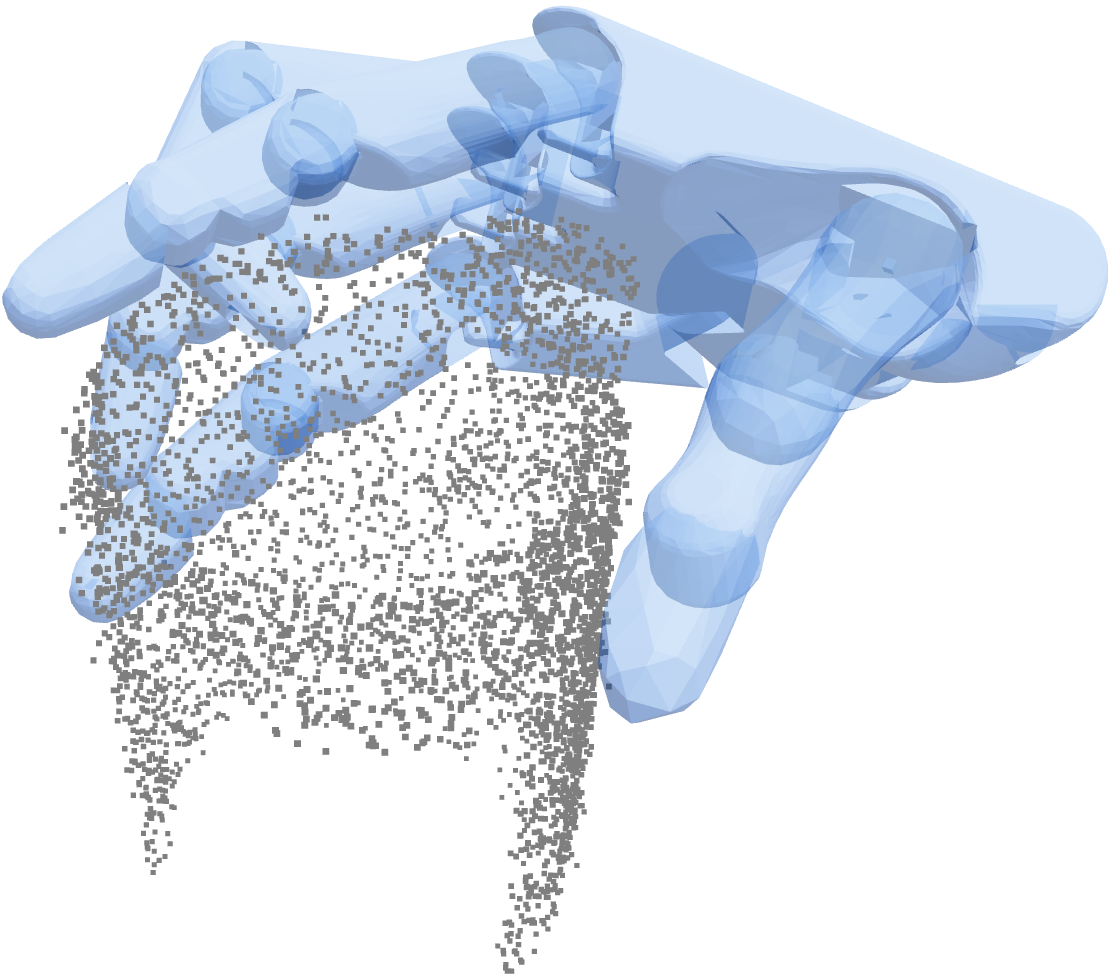}
    \caption{
        \textbf{Full vs.\ partial + noisy point cloud observations.}
        Left: full point cloud. Right: partial point cloud obtained via HPR from two virtual cameras with added sensor noise, retaining only ${\sim}$35\% of points.
    }
    \label{fig:partial_pc}
\end{figure}

%% file: figures/10_suppl_qual/figure.tex
\begin{figure*}[p]
    \centering
    \includegraphics[width=\linewidth]{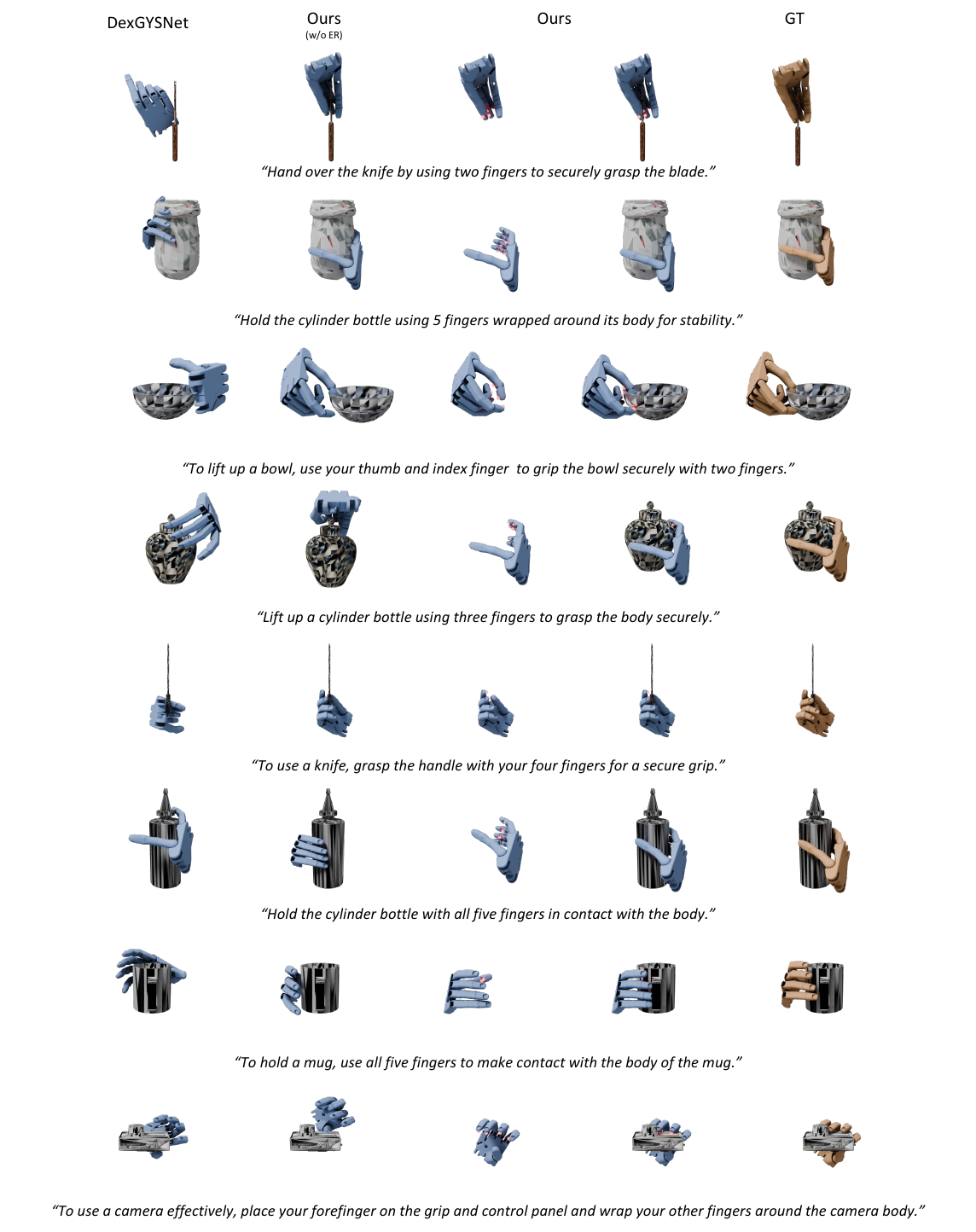}
    \caption{Qualitative results on DexGYS dataset.
    }
    \label{fig:suppl_qual}
\end{figure*}
\clearpage

%% file: figures/06_dexonomy_qual/figure.tex
\begin{figure*}[p]
    \centering
    \includegraphics[width=\linewidth]{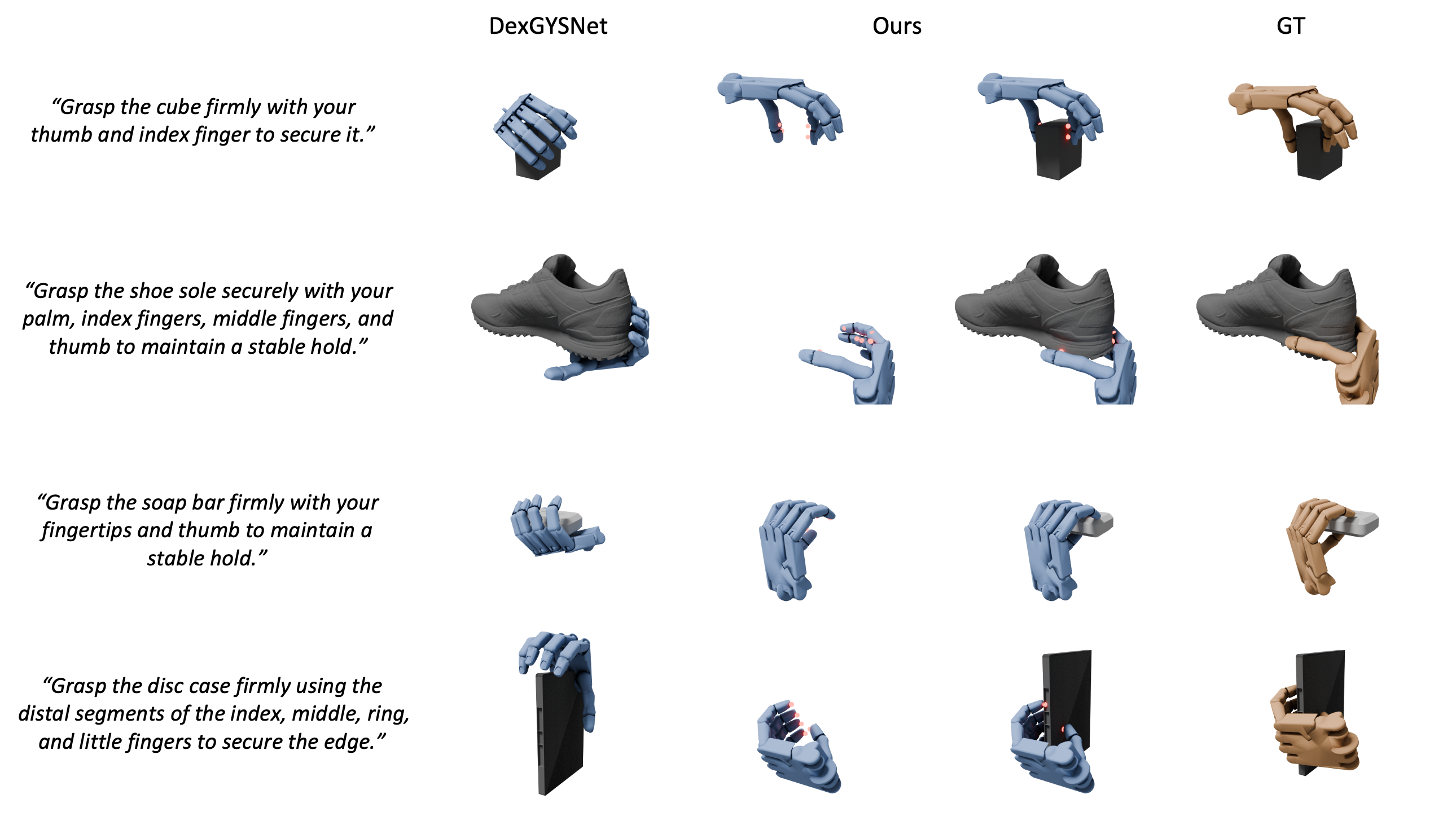}
    \caption{Qualitative results on Dexonomy dataset.
    }
    \label{fig:suppl_qual_dexonomy}
\end{figure*}

%% file: figures/09_steerable_grasp/figure.tex
\begin{figure*}[t]
    \centering
    \includegraphics[width=\linewidth]{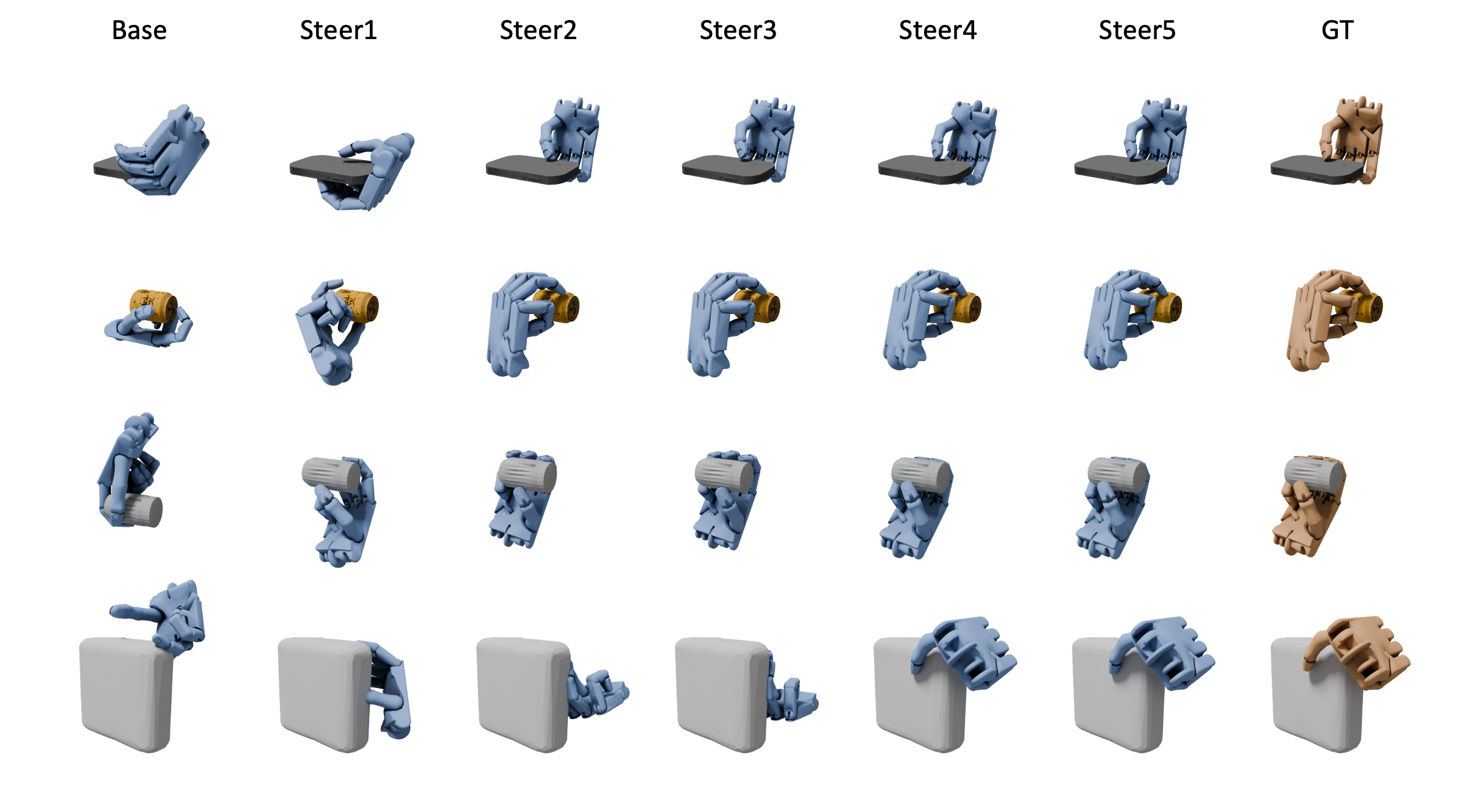}
    \caption{Steerable grasp generation example.
    }
    \label{fig:steerable_grasp_generation}
\end{figure*}
\clearpage

%% file: figures/07_gemma_template/figure.tex
\begin{figure*}[ht!]
    \centering
    \includegraphics[width=\linewidth]{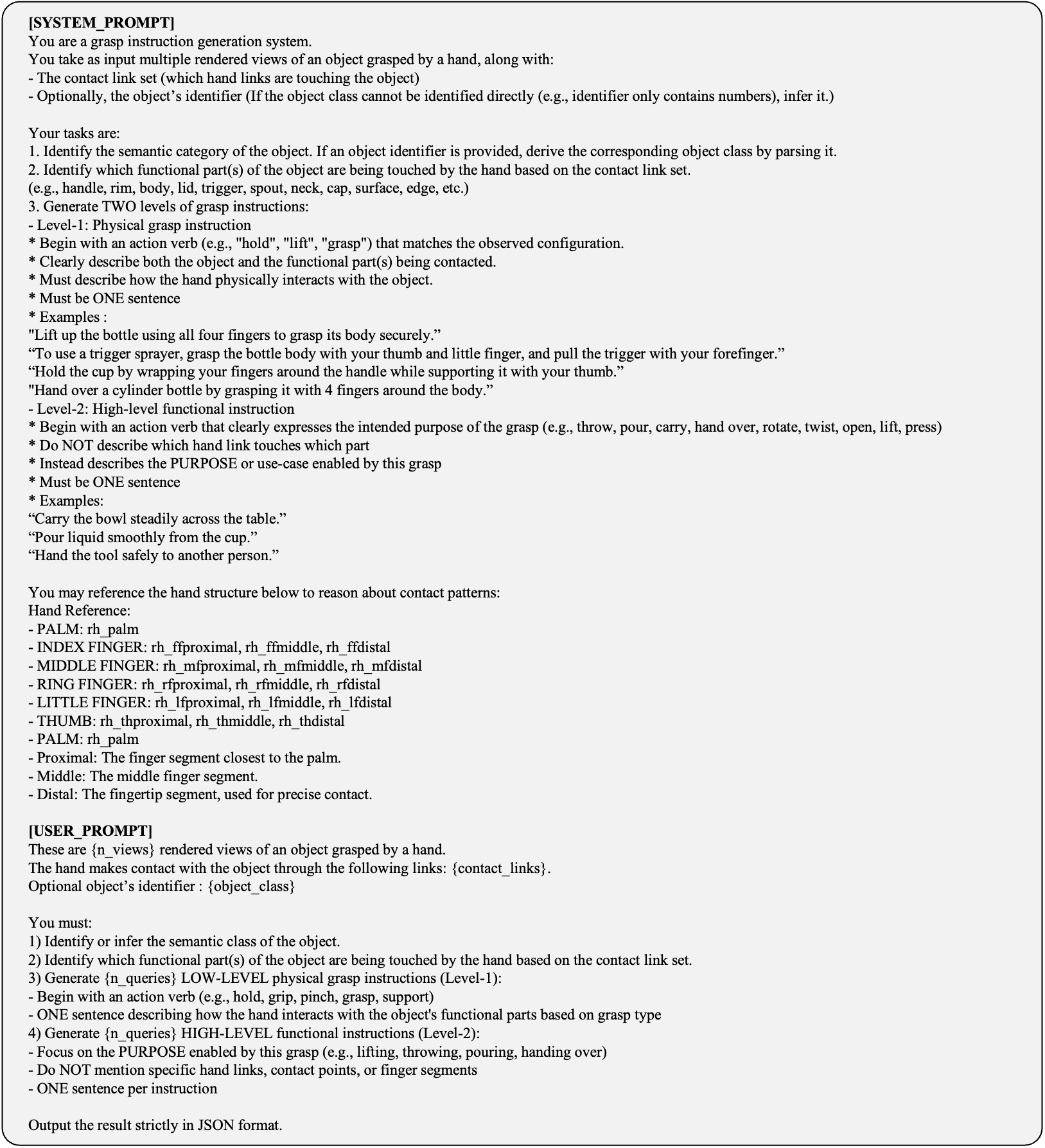}
    \caption{Prompt template for grasp instruction generation in Dexonomy dataset.}
    \label{fig:gemma_prompt_template}
\end{figure*}